\documentclass[lettersize,journal]{IEEEtran}
\usepackage{amsmath,amsfonts}
\usepackage{algorithmic}
\usepackage{algorithm}
\usepackage{array}
\usepackage[caption=false,font=normalsize,labelfont=sf,textfont=sf]{subfig}
\usepackage{textcomp}
\usepackage{stfloats}
\usepackage{url}
\usepackage{verbatim}
\usepackage{graphicx}
\usepackage{cite}
\usepackage{hyperref}
\hypersetup{hypertex=true,
	colorlinks=true,
	linkcolor=red,
	anchorcolor=red,
	citecolor=red}
\usepackage{booktabs}
\usepackage{makecell}
\usepackage{booktabs}
\usepackage{bbm}
\usepackage{multirow}

\usepackage{pifont}

\usepackage[capitalize]{cleveref}
\makeatother

\begin{document}
	
	\title{OAMatcher: An Overlapping Areas-based Network for Accurate Local Feature Matching}
	
	\author{Kun Dai$^{\dagger}$, Tao Xie$^{\dagger}$, Ke Wang, Zhiqiang Jiang, Ruifeng Li, Lijun Zhao
		
		\thanks{This work was in part by National Natural Science Foundation of China under Grant 62176072 and 62073101. (Corresponding author: Ruifeng Li and Ke Wang. ${\dagger}$: These authors contribute equally.)}
		\thanks{Kun Dai, Tao Xie, Ke Wang, Zhiqiang Jiang, Ruifeng Li, Lijun Zhao are with State Key Laboratory of Robotics and System, Harbin Institute of Technology, Harbin 150006, China (email: {20s108237@stu.hit.edu.cn;
				xietao1997@hit.edu.cn; 
				wangke@hit.edu.cn; 
				22s008043@stu.hit.edu.cn;
				lrf100@hit.edu.cn;
				zhaolj@hit.edu.cn).}}
		\thanks{Tao Xie is also with SenseTime Group Inc., Beijing 100080, China (email: xietao@sensetime.com).}}
	
	\markboth{Journal of \LaTeX\ Class Files,~Vol.~14, No.~8, August~2021}%
	{Shell \MakeLowercase{\textit{et al.}}: A Sample Article Using IEEEtran.cls for IEEE Journals}
	
	
	\maketitle
	
	\begin{abstract}
		Local feature matching is an essential component in many visual applications.
		In this work, we propose OAMatcher, a Tranformer-based detector-free method that imitates humans behavior to generate dense and accurate matches.
		Firstly, OAMatcher predicts overlapping areas to promote effective and clean global context aggregation, with the key insight that humans focus on the overlapping areas instead of the entire images after multiple observations when matching keypoints in image pairs.
		Technically, we first perform global information integration across all keypoints to imitate the humans behavior of observing the entire images at the beginning of feature matching.
		Then, we propose Overlapping Areas Prediction Module (OAPM) to capture the keypoints in co-visible regions and conduct feature enhancement among them to simulate that humans transit the focus regions from the entire images to overlapping regions, hence realizeing effective information exchange without the interference coming from the keypoints in non-overlapping areas.
		Besides, since humans tend to leverage probability to determine whether the match labels are correct or not, we propose a Match Labels Weight Strategy (MLWS) to generate the coefficients used to appraise the reliability of the ground-truth match labels, while alleviating the influence of measurement noise coming from the data.
		Moreover, we integrate depth-wise convolution into Tranformer encoder layers to ensure OAMatcher extracts local and global feature representation concurrently.
		Comprehensive experiments demonstrate that OAMatcher outperforms the state-of-the-art methods on several benchmarks, while exhibiting excellent robustness to extreme appearance variants.
		The source code is available at https://github.com/DK-HU/OAMatcher.
		
	\end{abstract}
	
	\begin{IEEEkeywords}
		Local Feature Matching, Transformer, Overlapping Areas Prediction.
	\end{IEEEkeywords}

	\section{Introduction}
	\IEEEPARstart{L}{ocal} feature matching is a vital component in several visual applications, such as Simultaneous Localization and Mapping (SLAM) \cite{mur2017orb, campos2021orb, qin2018vins} and Structure-from-Motion (SFM) \cite{schonberger2016structure, cui2022vidsfm}.
	
	Broadly, the local feature matching methods can be divided into detector-based and detector-free methods.
	The conventional detector-based approaches \cite{zhao2022alike, fan2022seeing, lowe2004distinctive, rublee2011orb, DeTone_2018_CVPR_Workshops, Dusmanu_2019_CVPR, revaud2019r2d2, tyszkiewicz2020disk, sarlin2020superglue, chen2021learning, kuang2021densegap, shi2022clustergnn, detone2018superpoint, luo2020aslfeat} first extract a set of keypoints, utilize handcrafted \cite{rublee2011orb, lowe2004distinctive, bay2006surf} or learning-based high-dimension vectors \cite{DeTone_2018_CVPR_Workshops, Dusmanu_2019_CVPR, revaud2019r2d2, tyszkiewicz2020disk, zhao2022alike} to describe them, and then implements matching algorithm \cite{tao2002continuous, bian2017gms, zhang2019learning, chen2022csr} to generate correspondences.
	Being extensively investigated, the detector-based methods achieve tremendous success in ideal environments, while they suffer from severe performance degeneration when handling image pairs with extreme appearance variants since the detectors struggle to extract repeatable keypoints.
	
	\begin{figure}[t]
		\centering
		\includegraphics[width=0.999\hsize]{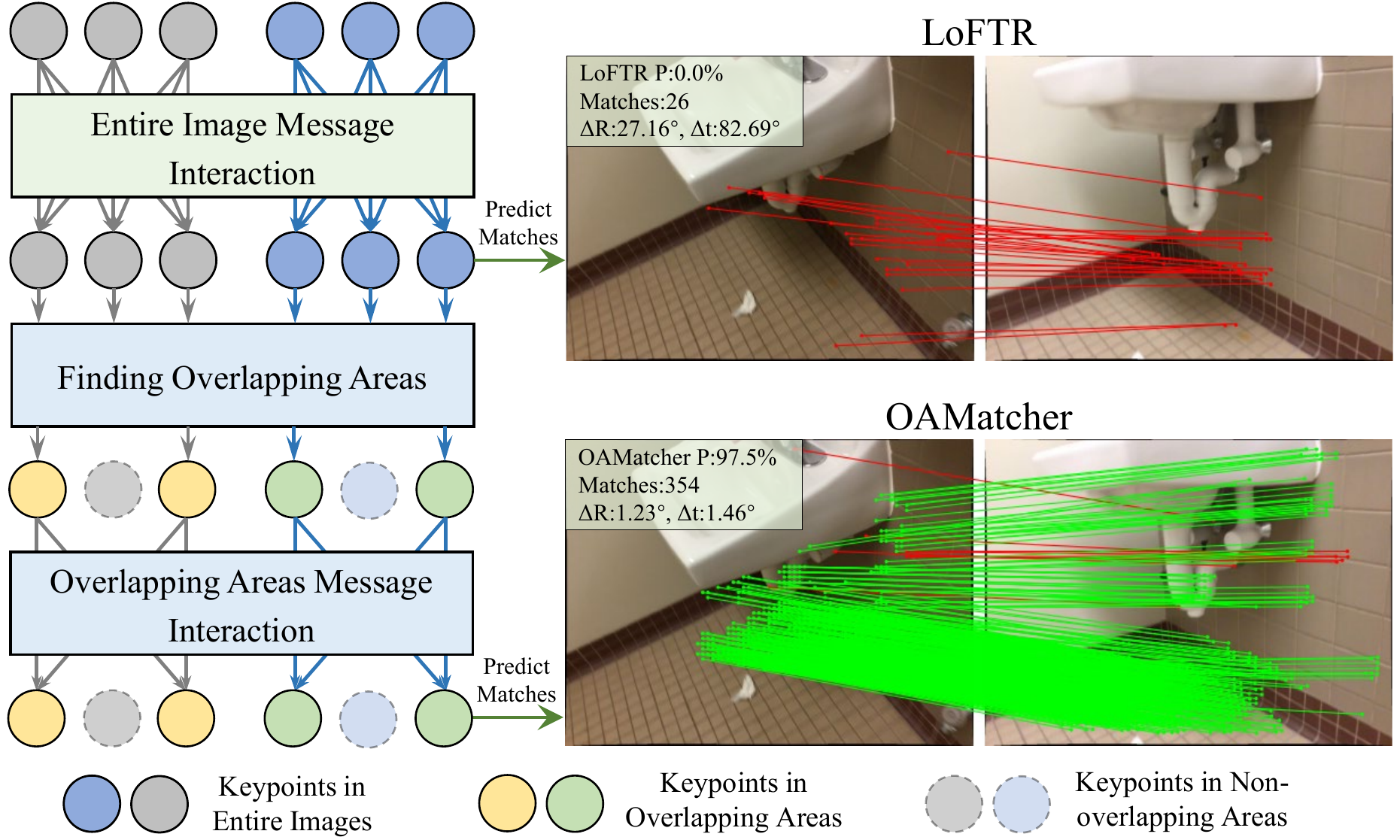}
		\caption{\textbf{Comparison between LoFTR and OAMatcher.} Compared with the LoFTR that only integrates information in the entire images, OAMatcher transits the focus regions from entire images to overlapping regions, which is more human-intuitive.
		}
		\label{covision_process}
		\vspace{-10pt}
	\end{figure}
	Concurrent with the detector-based matching methods, the detector-free approaches \cite{ma2022correspondence, rocco2018neighbourhood, rocco2020efficient, li2020dual, zhou2021patch2pix, efe2021dfm, sun2021loftr, wang2022matchformer, truong2022topicfm, chen2022aspanformer, efe2021dfm, sun2019guide} focus on extracting visual descriptors on dense grids across images directly, therefore, the repeatable keypoints in image pairs can be captured \cite{sun2021loftr}.
	Conventional detector-free methods \cite{rocco2018neighbourhood, rocco2020efficient, li2020dual, zhou2021patch2pix, efe2021dfm} utilize CNNs to extract visual descriptors, while they cannot differentiate keypoints with similar feature representation owing to the limited receptive field of CNNs.
	Recently, Transformer \cite{vaswani2017attention} has been successfully applied for establishing precise matches due to its powerful capability of modelling long-range global context information \cite{sarlin2020superglue, sun2021loftr}. 
	As a pioneering work, LoFTR \cite{sun2021loftr} views keypoints as nodes to construct a graphy neural network (GNN), in which the Linear Transformer \cite{katharopoulos2020transformers} are leveraged to aggregate global message intra/inter images.
	While detector-free methods deliver extraordinary results, they come with some issues:
	
	(i) Existing detector-free methods consistently integrate the features of all keypoints without changing focus regions, which is not in line with humans behavior.
	Intuitively, humans observe the entire images at the begining of feature matching, and then focus on the overlapping areas after multiple observations.
	Without this process, the keypoints in non-overlapping areas will disturb us to select correct matches.
	Therefore, extracting the keypoints in co-visible regions is conducive for network to perform effective and clean context information aggregation.
	(ii) Owing to the measurement noise coming from data (e.g, the values of depth maps are inaccurate), humans perfer to leverage probability to determine whether the match labels are reliable or not, hence alleviating the influence of the unreliable matches labels, as shown in \cref{MLWS_fig}.
	However, existing detector-free methods only utilize $0$ or $1$ to appraise the match labels, which poses an ineluctable obstacle for training process.
	

	To rememdy the above problems, we propose OAMatcher, a detector-free architecture that generates dense and accurate matches in a human-intuitive manner.
	For the first issue, we effectively predict co-visible masks to imitate humans behavior of looking from entire images to overlapping regions, hence ensuring efficient and clean message exchange. 
	Specifically, we first utilize self/cross attention layers to propagate the global context information across all keypoints intra/inter images.
	Then, we propose the \textbf{Overlapping Areas Prediction Module (OAPM)} to capture the keypoints in co-visible regions.
	Concretely, we compute the soft assignment matrix of all keypoints and select the maximum along row and column dimension to derive probability maps, which are handled by an adaptive threshold for binarization.
	Subsequently, we leverage morphological close operation to connect the adjacent mask regions and fill the pixels within its maximum contours to generate co-visible masks, hence capturing the keypoints in overlapping areas.
	Ultimately, we interleave self/cross attention layers among the selected keypoints to effectively integrate the context information without the interference coming from the keypoints in non-overlapping areas.
	By performing the above procedures, the focus regions of OAMatcher are transited from entire images to overlapping areas.
	
	For the second issue, we propose a \textbf{Match Labels Weight Strategy (MLWS)} to imitate the humans behavior of using probability to appraise whether the matches labels are reliable or not, while alleviating measurement noise coming from the data.
	Specifically, we wrap the keypoints in the first image to the second image to obtain projection points, which locate in grids that utilize four keypoints as corners.
	Then, we calculate the Euclidean distance between the projection points and four corners of the grids, from which the two corners with less distance are selected to make label confidence.
	Ultimately, we perceive the label confidence as the coefficients used to weight the loss function, hence ensuring the match labels with high confidence better supervise the network training.
	
	To summarize, the main contributions of this work are as follows:
	
	\begin{itemize}
		\item{We propose OAMatcher, a detector-free network that generates accurate matches in a human-intuitive manner.}
		\item{We propose an Overlapping Areas Prediction Module that captures the co-visible regions of image pairs to ensure effective and clean context information aggregation.}
		\item{We propose a Match Labels Weight Strategy that utilizes label confidence to weight loss function, hence relieving the influence of the inaccurate match labels.}
		\item{We demonstrate that OAMatcher outperforms the cutting-edge methods on several benchmarks, while exhibiting excellent robustness to extreme appearance variants.}
	\end{itemize} 
	
	The rest of this paper is organized as follows.
	In \cref{S2}, we give a brief introduction about related works.
	In \cref{S3}, we elaborate the proposed method OAMatcher in detail.
	In \cref{S4}, we conduct experiments on multiple tasks. 
	Ultimately, a conclusion is given in \cref{S5}.
	
	\section{Related Work} \label{S2}
	\subsection{Detector-based Methods}
	The conventional detector-based methods \cite{lowe2004distinctive, rublee2011orb, DeTone_2018_CVPR_Workshops, Dusmanu_2019_CVPR, revaud2019r2d2, tyszkiewicz2020disk, sarlin2020superglue, chen2021learning, kuang2021densegap, shi2022clustergnn, detone2018superpoint, luo2020aslfeat} are realized by (i) detecting and describing the sparse keypoints with elaborate detectors \cite{lowe2004distinctive, rublee2011orb, karpushin2016keypoint, DeTone_2018_CVPR_Workshops, Dusmanu_2019_CVPR, revaud2019r2d2, tyszkiewicz2020disk, luo2020aslfeat}. (ii) establishing pixel-wise correspondences by nearest neighbor search or more sophisticated matching algorithms \cite{bian2017gms, zhang2019learning}.
	With the development of deep learning, substantial methods leverage CNNs to extract keypoints with more robust and discriminative descriptors, such as D2-Net \cite{Dusmanu_2019_CVPR}, R2D2 \cite{revaud2019r2d2}, SuperPoint \cite{detone2018superpoint}, ASLFeat \cite{luo2020aslfeat}, and ALIKE \cite{zhao2022alike}.
	However, the limited receptive field of CNNs is the major impediment for these methods to distinguish keypoints with similar feature representation.
	
	Recently, Transformer \cite{vaswani2017attention} has attracted considerable interest in computer vision owing to its excellent capability of capturing long-range relationships \cite{li2022exploiting, jiayao2022real, pei2022transformer, fu2018refinet, song20166}.
	Therefore, the cutting-edge feature matching approaches \cite{sarlin2020superglue, chen2021learning, kuang2021densegap, shi2022clustergnn} perceive the keypoints as nodes to construct GNN, in which the Transformer are utilized to integrate global context information intra/inter images.
	As a pioneering work, SuperGlue \cite{sarlin2020superglue} establishes complete GNN architecture over keypoints and utilize self/cross attention to update their feature representation by exchanging global visual and gemetric messages.
	Then, the Sinkhorn algorithm \cite{cuturi2013sinkhorn} is utilized to generate precise matches.
	Although exhibiting excellent matching performance, SuperGlue consumes tremendous memory footprints since the matrix multiplication in Vanilla Transformer \cite{vaswani2017attention} leads to quadratic computational complexity with respect to the number of keypoints.
	Therefore, many methods try to optimize the structure of SuperGlue.
	SGMNet \cite{chen2021learning} introduces seeds as bottlenecks to construct sparse GNN, hence reducing the dominating computational complexity of the attention layers.
	ClusterGNN \cite{shi2022clustergnn} employs a progressive clustering module to divide the keypoints of all images into different subgraphs to reduce computation.
	Nevertheless, being detector-based methods, they have the fundamental drawback of being unable to detect repeatable keypoints when handling image pairs with extreme appearance changes \cite{sun2021loftr}.
	
	\subsection{Detector-free Methods}
	Instead of leveraging elaborate detectors to extract keypoints with descriptors, the detector-free methods \cite{rocco2018neighbourhood, rocco2020efficient, li2020dual, zhou2021patch2pix, efe2021dfm, sun2021loftr, wang2022matchformer, truong2022topicfm, chen2022aspanformer, efe2021dfm} remove the feature detector phase and directly produce descriptors on dense grids across images.
	Earlier methods \cite{rocco2018neighbourhood, rocco2020efficient, li2020dual, zhou2021patch2pix, efe2021dfm} generally utilize CNNs based on correlation or cost volume to identify probable neighbourhood consensus. 
	DRC-Net \cite{li2020dual} leverages coarse features to produce coarse 4D correlation tensor, which are utilized to guide the fine features to generate final dense correspondences.
	DFM \cite{efe2021dfm} extracts features with the pre-trained off-the-shelf VGG19 \cite{simonyan2014very}, aligns the image pair using the predicted homography matrix, and then applies hierarchical refinement to realize accurate image matches.
	
	Upon witnessing the great success of SuperGlue \cite{sarlin2020superglue}, LoFTR \cite{sun2021loftr} innovatively design a Transformer-based detector-free architecture, achieving outstanding matching performance.
	Technically, LoFTR utilizes Linear Transformer \cite{katharopoulos2020transformers} to promote the representation ability of visual descriptors by aggregating global context information.
	Subsequently, LoFTR calculates soft assignment matrix to capture coarse matches, which are optimized to fine matches with a correlation-based refinement block.
	Since then, copious detector-free approaches are conducted on the basis of LoFTR. 
	Matchformer \cite{wang2022matchformer} proposes a novel extract-and-match scheme that interleaves self and cross attention in each stage to perform feature extraction and feature similarity learning synchronously.
	QuadTree \cite{tang2022quadtree} introduces a quadtree attention that constructs token pyramids and performs message aggregation in a coarse-to-fine manner.
	TopicFM \cite{truong2022topicfm} divides the same spatial structure of image pairs into a same topic and augments the features inside each topic for accurate matching. 
	ASpanFormer \cite{chen2022aspanformer} regresses the flow maps in each cross-attention phase to perform local attention.
	However, existing Transformer-based detector-free methods dose not shift the focus regions of networks and ignore the influence of the unreliable match labels caused by measurement noise.
	Therefore, prompting the network to focus on the co-visible regions and relieving the influence of the improper labels warrant further exploring.
	\begin{figure*}
		\centering
		\includegraphics[width=0.99\hsize]{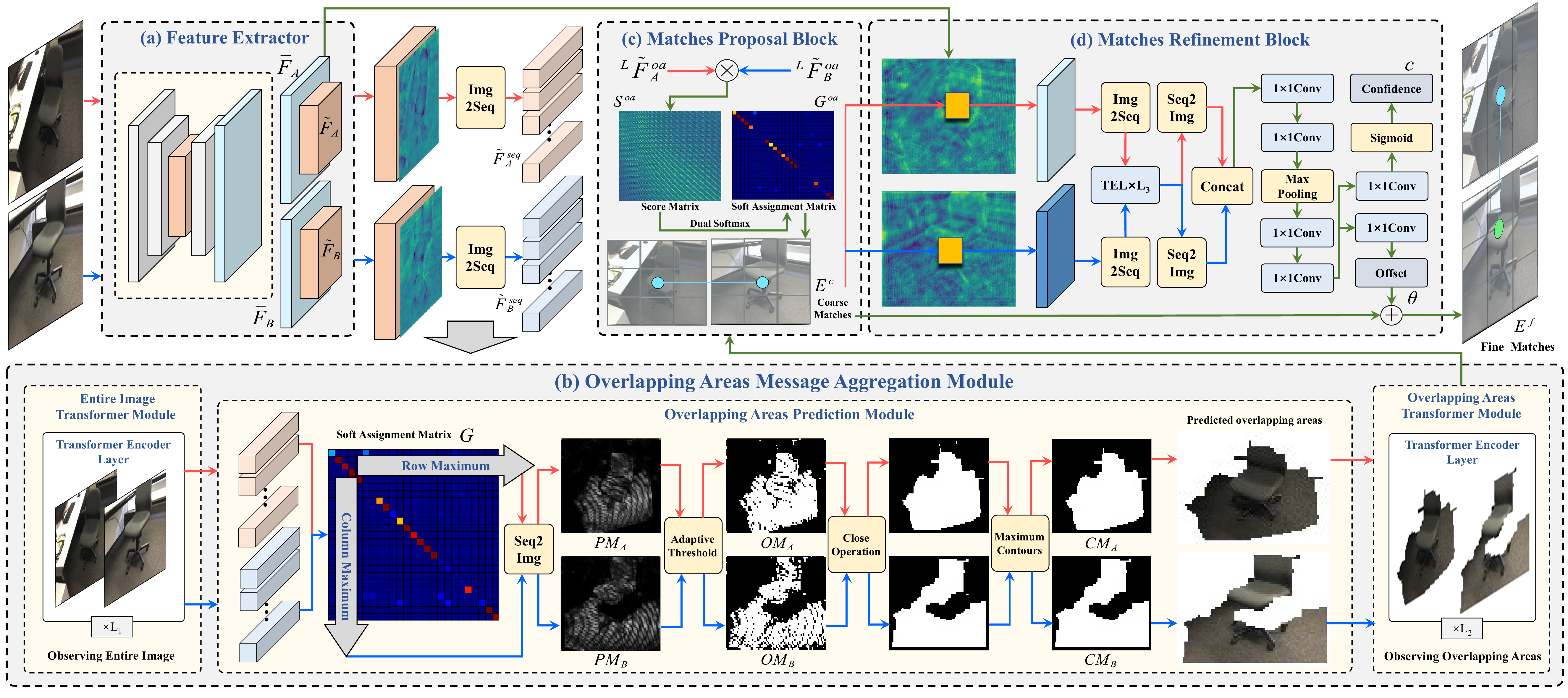}
		\caption{\textbf{The network architecture of OAMatcher.} OAMatcher utilizes \textbf{Feature Extractor} to generate multi-scale features. Then, OAMatcher leverages \textbf{Overlapping Areas Message Aggregation Module} to capture co-visible regions, realizing effective and clean context message passing. Finally, \textbf{Matches Proposal Block} are proposed to predict coarse matches, which are optimized by \textbf{Matches Refinement Block} to generate final fine matches.
		}
		\label{overall}
	\end{figure*}
	\subsection{Efficient Transformer Architecture}
	In the vanilla Transformer, the memory consumption is quadratic with respect to the length of sequences.
	To rememdy this problem, many works are proposed to sparsify connections or linearize the attention complexity of Transformer.
	Beltagy et al. \cite{beltagy2020longformer} compute a sparse attention matrix to cut down computational cost.
	Katharopoulos et al. \cite{katharopoulos2020transformers} utilize the associativity property of matrix products to realize linear computational complexity.
	Wu et al. \cite{wu2021fastformer} leverage element-wise product to obtain global context-aware attention value, which is combined with the query vector to form the final results.
	In this work, we follow LoFTR and utilize Linear Transformer to control the computation of networks.

	\section{Methodology} \label{S3}
	\subsection{Overall}
	Intuitively, humans first observe the entire images when performing feature matching and then shift the focus regions to the overlapping areas.
	Besides, we prefer to use probability to appraise whether the match labels are reliable or not.
	Consequently, as shown in \cref{overall}, we propose a Transformer-based detector-free method OAMatcher to imitate these behaviors.
	
	Technically, OAMatcher utilizes a CNN-based encoder to extract fine-level and coarse-level features concurrently and perceives the pixels of coarse-level features as keypoints.
	Then, we propose an \textbf{Overlapping Areas Message Aggregation Modules (OMAM)} that first performs feature enhancement among all keypoints to imitate humans behavior of integrating information from the entire images at the begining of feature matching.
	Subsequently, we introduce an \textbf{Overlapping Areas Prediction Module (OAPM)} to capture the keypoints in co-visible regions and then propagate the long-range context information among the selected keypoints so as to simulate that humans focus on the overlapping areas after multiple observations.
	Ultimately, we utilize a \textbf{Matches Proposal Block (MPB)} to predict coarse matches, which are refined by \textbf{Matches Refinement Block (MRB)} to generate final matches.
	
	Moreover, we propose a \textbf{Match Labels Weight Strategy (MLWS)} to generate label confidence used to weight loss function, thus relieving the interference of the measurement noise coming from the data.
	In the following part, we introduce the details and underlying insights of each individual block.
	
	\subsection{Feature Extractor}
	In this stage, we utilize ResNet18 \cite{he2016deep} with FPN \cite{lin2017feature} to extract fine-level features $\bar{F}_{A}, \bar{F}_{B} \in \mathbb{R}^{\bar{C} \times H/2 \times W/2}$ and coarse-level features $\tilde{F}_{A}, \tilde{F}_{B} \in \mathbb{R}^{\tilde{C} \times H/8 \times W/8}$, where $H, W$ mean the height and width of images, $\bar{C}, \tilde{C}$ denote feature dimension.
	For convenience, we define $N = H/8 \times W/8$.
	Since each pixel in coarse-level features represents an $8 \times 8$ grid in original images $I_{A}, I_{B}$, we perceive the central pixels of grids as keypoints $P_{A}, P_{B} \in \mathbb{R}^{N \times 2}$ with the corresponding features in $\tilde{F}_{A}, \tilde{F}_{B}$ as visual descriptors.

	\subsection{Overlapping Areas Message Aggregation Module (OMAM)}
	
	\textbf{Transformer Encoder Layer (TEL).}
	Firstly, we briefly introduce the Transformer Encoder Layer (TEL), which is comprised of Linear Attention Layer (LAL) and Feed-forward Network (FFN).
	For LAL, we follow LoFTR \cite{sun2021loftr} and utilize self/cross linear attention \cite{katharopoulos2020transformers} to perform long-range global context aggregation intra/inter images.
	For self-attention, the input features $U$ and $R$ are same. 
	For cross-attention, the input features $U$ and $R$ are different.
	LAL transforms the input features $U, R$ into the query, key, and value vectors $Q,K,V$:
	\begin{equation}
		\begin{split}
			Q = U W_Q, \ \ \ K = R W_K, \ \ \ V = R W_V,
		\end{split}
	\end{equation}
	where $W_Q, W_K, W_V$ are learnable parameters.
	Subsequently, the linear attention layer is formatted as:
	\begin{equation}
		M = \phi(Q)(\phi(K)^{T}V),
	\end{equation}
	where $\phi(\cdot) = elu(\cdot) + 1$,  $elu(\cdot)$ is the exponential linear unit activation function.
	We define the process of LAL as $M = LAL(U,R)$.
	
	The Transformer Encoder Layer of other works \cite{sarlin2020superglue, sun2021loftr, chen2021learning, shi2022clustergnn} only focuses on propagating global context information, while ignoring extracting local feature representations.
	As a consequence, the local features are discarded constantly with the Transformer Encoder Layer goes deeper.
	Therefore, we integrate depth-wise convolution \cite{howard2017mobilenets} into FFN to extract local and global information concurrently.
	Specifically, we concatenate the input feature $U$ and the global message $M$ along the channel dimension and expand the channel dimension by $\gamma / 2$ times with a MLP to extract more abundant features.
	Then, we convert the sequences to 2D feature maps to recover the geometry relationship between keypoints, followed by a depth-wise convolution with kernel size of $3$ to extract local information.
	Subsequently, we flatten the feature maps to sequences and utilize a GELU activation function and a MLP to generate enhanced features with both global and local information.
	This process can be formulated as:
	\begin{equation}
		\begin{aligned}
			FFN(U, M) &= U + MLP_{1/\gamma}(GELU(Img2Seq(\\
			&DW(Seq2Img(MLP_{\gamma/2}([U||M])))))),
		\end{aligned}
	\end{equation}
	where $MLP_{1/\gamma}, MLP_{\gamma/2}$ mean expand the channel dimension by $1/\gamma, \gamma/2$ times with a MLP, respectively;
	$[\cdot||\cdot]$ means concatenation along the channel dimension;
	$Seq2Img(\cdot)$ converts sequences to 2D feature maps;
	$DW(\cdot)$ means depth-wise convolution;
	$Img2Seq$ converts 2D feature maps to sequences;
	$GELU(\cdot)$ means GELU activation function.
	
	In summary, we format the Transformer Encoder Layer as: 
	\begin{equation}
		TEL(U,R) = FFN(U, LAL(U,R))
	\end{equation}
	
	\textbf{Entire Image Transformer Module (EITM).}
	Firstly, we integrate the information across all keypoints $P_{A}, P_{B}$ to imitate the humans behavior of observing entire images at the beginning of feature matching.
	Specifically, we flatten the coarse-level features $\tilde{F}_{A}, \tilde{F}_{B}$ into sequences $\tilde{F}_{A}^{seq}, \tilde{F}_{B}^{seq} \in \mathbb{R}^{N \times \tilde{C}}$, and then interleave the Transformer Encoder Layer by $L_{1}$ times to enhance features.
	For the $l$-th feature enhancement, we utilize self attenion ($U, R$ are same, i.e. either $(\tilde{F}_{A}^{seq}, \tilde{F}_{A}^{seq})$ or ($\tilde{F}_{B}^{seq}, \tilde{F}_{B}^{seq}$)) and cross attention ($U, R$ are different, i.e. either $(\tilde{F}_{A}^{seq}, \tilde{F}_{B}^{seq})$ or ($\tilde{F}_{B}^{seq}, \tilde{F}_{A}^{seq}$)) to exchange context information intra/inter images, which can be formatted as:
	\begin{equation}
		\begin{aligned}
			^{l-1}\tilde{F}^{seq}_{A} &= TEL(^{l-1}\tilde{F}^{seq}_{A},\ ^{l-1}\tilde{F}^{seq}_{A}), \\
			^{l-1}\tilde{F}^{seq}_{B} &= TEL(^{l-1}\tilde{F}^{seq}_{B},\ ^{l-1}\tilde{F}^{seq}_{B}), \\
			^{l}\tilde{F}^{seq}_{A} &= TEL(^{l-1}\tilde{F}^{seq}_{A},\ ^{l-1}\tilde{F}^{seq}_{B}), \\
			^{l}\tilde{F}^{seq}_{B} &= TEL(^{l-1}\tilde{F}^{seq}_{B},\ ^{l}\tilde{F}^{seq}_{A})
		\end{aligned}
	\end{equation}
	
	Ultimately, we integrate the local and global information into discriminative features $^{L}\tilde{F}^{seq}_{A}, ^{L}\tilde{F}^{seq}_{B} \in \mathbb{R}^{N \times \tilde{C}}$.
	
	\textbf{Overlapping Areas Prediction Module (OAPM).}
	After the information aggregation across entire images, we propose Overlapping Areas Prediction Module (OAPM) to capture the overlapping regions of image pairs, hence imitating humans behavior of looking from entire images to co-visible areas after multiple observations.
	Technically, we first utilize inner product of $^{L}\tilde{F}^{seq}_{A}, ^{L}\tilde{F}^{seq}_{B}$ to calculate the score matrix $S \in \mathbb{R}^{N \times N}$, followed by a dual-softmax algorithm to generate soft assignment matrix $G \in \mathbb{R}^{N \times N}$:
	\begin{equation}
		\begin{aligned}
			S(i,j) &= \langle ^{L}\tilde{F}^{seq}_{A}(i), ^{L}\tilde{F}^{seq}_{B}(j) \rangle, \\
			G = Soft&max(S)_{row} \odot Softmax(S)_{col},
		\end{aligned}
		\label{score_assign}
	\end{equation}
	where $\langle \cdot , \cdot \rangle$ denotes the inner product;  
	$Softmax(\cdot)_{row}$, $\ Softmax(\cdot)_{col}$ mean performing softmax on each row and column of $S$;
	$\odot$ means element-wise multiplication.
	
	Since each row/column of $G$ represents the response values of keypoints in the first/second image to keypoints in the second/first image, we consider that the keypoints with large response values are more likely to locate in overlapping areas.
	Therefore, we calculate the maximum of $G$ along the row and column and convert their into 2D images to derive probability maps $PM_{A}, PM_{B} \in \mathbb{R}^{H/8 \times W/8}$, which can be interpreted as the probability estimates of whether the keypoints locate in the co-visible regions:
	
	\begin{equation}
		\begin{aligned}
			PM_{A} &= Seq2Img(Max(G)_{row}), \\ 
			PM_{B} &= Seq2Img(Max(G)_{col}),
		\end{aligned}
	\end{equation}
	where $Max(\cdot)_{row}, Max(\cdot)_{col}$ mean calculaing the maximum along row and column dimension;
	$Seq2Img(\cdot)$ means converting sequences to 2D feature maps.
	
	\begin{figure}
		\centering
		\includegraphics[width=0.999\hsize]{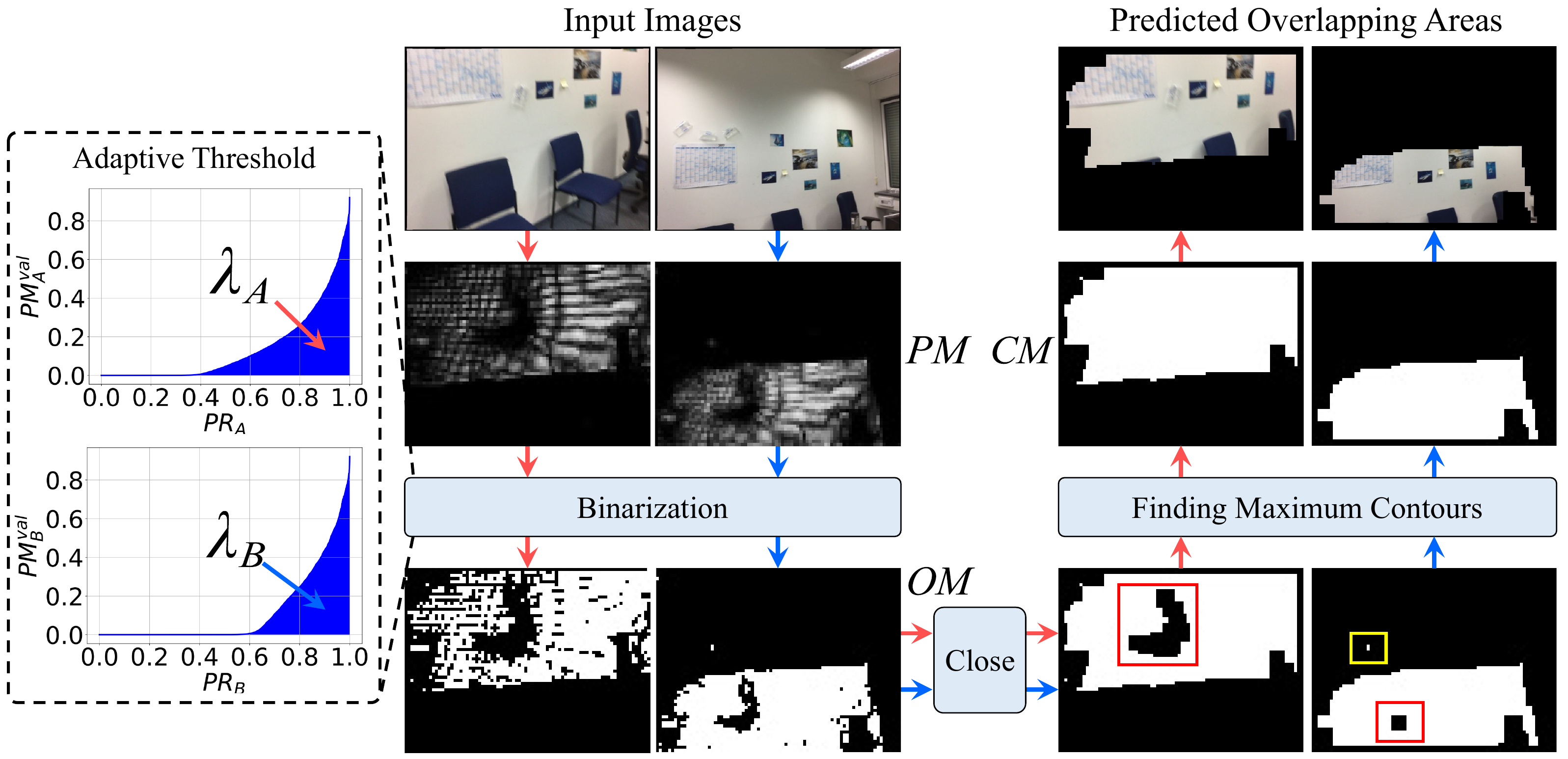}
		\caption{\textbf{The illustration of OAPM.} OAPM utilizes adaptive threshold, morphological close operation and maximum contours to generate overlapping areas.
		}
		\label{covision_process}
		\vspace{-10pt}
	\end{figure}

	Subsequently, we utilize the following three procedures to capture overlapping areas.
	
	(i) Binarization. To obtain the co-visible masks, we first set a threshold to binarize the probability maps $PM_{A}, PM_{B}$.
	However, it is irrational to set a fixed threshold since the probability map values of different image pairs are significantly inconsistent.
	To overcome this coundrum, we propose an adaptive threshold that can be adjusted flexibly according to the image pairs.
	As shown in \cref{covision_process}, take the $PM_{A}$ as an example, we first sort the values of $PM_{A}$ in an ascending order, obtaining $PM_{A}^{val} \in \mathbb{R}^{N}$.
	Then, we divide a series of proportions ranging from $0$ to $1$ according to the length of the sequence $PM_{A}^{val}$, deriving $PR_{A} \in \mathbb{R}^{N}$:
	\begin{equation}
		PR_{A} = \{ \frac{k}{N} \ | \ k=0,1,...,N-1 \}
	\end{equation}
	
	Subsequently, as illustrated in \cref{covision_process}, we plot a curve with $PR_{A}, PM_{A}^{val}$ as $X, Y$ axes and calculate the area under the curve as the adaptive threshold $\lambda_{A}$. 
	Given $\lambda_{A}$, $\lambda_{B}$, the probability maps $PM_{A}, PM_{B}$ are binarized to generate the original  co-visible mask $OM_{A}, OM_{B}$.
	
	(ii) Morphological Close Operation: Since the overlapping areas of the image pairs are connected, we leverage morphological close operation with kernel size of $k$ to process $OM_{A}, OM_{B}$, hence connecting the adjacent mask areas.
	
	(iii) Finding Maximum Contours: To eliminate the holes (red rectangles marked in \cref{covision_process}) in the predicted masks and alleviate the interference coming from the noise (yellow rectangle marked in \cref{covision_process}), we seek the maximum contours of $OM_{A}, OM_{B}$ and fill the pixels within them to generate the final co-visible masks $CM_{A}, CM_{B} \in \mathbb{R}^{H/8 \times W/8}$.
	
	In accordance with the co-visible masks $CM_{A}, CM_{B}$ and the visual descriptors $^{L}\tilde{F}^{seq}_{A}, ^{L}\tilde{F}^{seq}_{B}$ of keypoints, we can extract $K$ co-visible keypoints $P_{A}^{oa}, P_{B}^{oa} \in \mathbb{R}^{K \times 2}$ in overlapping areas with the descriptors defined as $\tilde{F}_{A}^{oa}, \tilde{F}_{B}^{oa} \in \mathbb{R}^{K \times \tilde{C}}$.

	\textbf{Overlapping Areas Transformer Module (OATM).}
	At this stage, we perform feature enhancement among the keypoints in co-visible areas without the interference coming from the keypoints in non-overlapping regions.
	Similar to EITM, OATM interleaves the Transformer Encoder Layer by $L_{2}$ times for features enhancement, generating the enhanced visual descriptors $^{L}\tilde{F}_{A}^{oa}, ^{L}\tilde{F}_{B}^{oa} \in \mathbb{R}^{K \times \tilde{C}}$.
	Notably, the depth-wise convolution in Transformer Encoder Layers is discarded since we cannot convert $\tilde{F}_{A}^{oa}, \tilde{F}_{B}^{oa}$ to 2D feature maps.
	
	\subsection{Matches Proposal Block (MPB)}
	Similar to \cref{score_assign}, we utilize $^{L}\tilde{F}_{A}^{oa}, ^{L}\tilde{F}_{B}^{oa}$ to calculate the co-visible score matrix $S^{oa} \in \mathbb{R}^{K \times K}$ and co-visible soft assignment matrix $G^{oa} \in \mathbb{R}^{K \times K}$.
	Then, for the $i$-th keypoint in the first image and the $j$-th keypoint in the second image, we perceive them as a coarse match if they satisfy the following conditions:
	(i) The soft assignment matrix values are larger than a predefined threshold $\rho$: $G^{oa}(i,j) > \rho$,
	(ii) They satisfy the mutual nearest neighbor (MNN) criteria, i.e., $G^{oa}(i,j)$ is the maximum in the $i$-th row and the $j$-th colume of $G^{oa}$.
	Ultimately, we derive the index $D$ of the keypoints in coarse matches:
	\begin{equation}
		D = \{ (i,j) \ | \ \forall \ G^{oa}(i,j) > \rho, \ (i,j) \in MNN(G^{oa})\}
	\end{equation}
	
	Given the keypoints $P_{A}^{oa}, P_{B}^{oa}$ and the index $D$, we define the coarse matches $E^{c} = \{ (P_{A}^{c}, P_{B}^{c}) \}$ as:
	\begin{equation}
		E^{c} = \{ (P_{A}^{oa}(i), P_{B}^{oa}(j)) \ | \ \forall(i,j) \in D \}
  	\end{equation}
	
	\begin{figure}
		\centering
		\includegraphics[width=0.999\hsize]{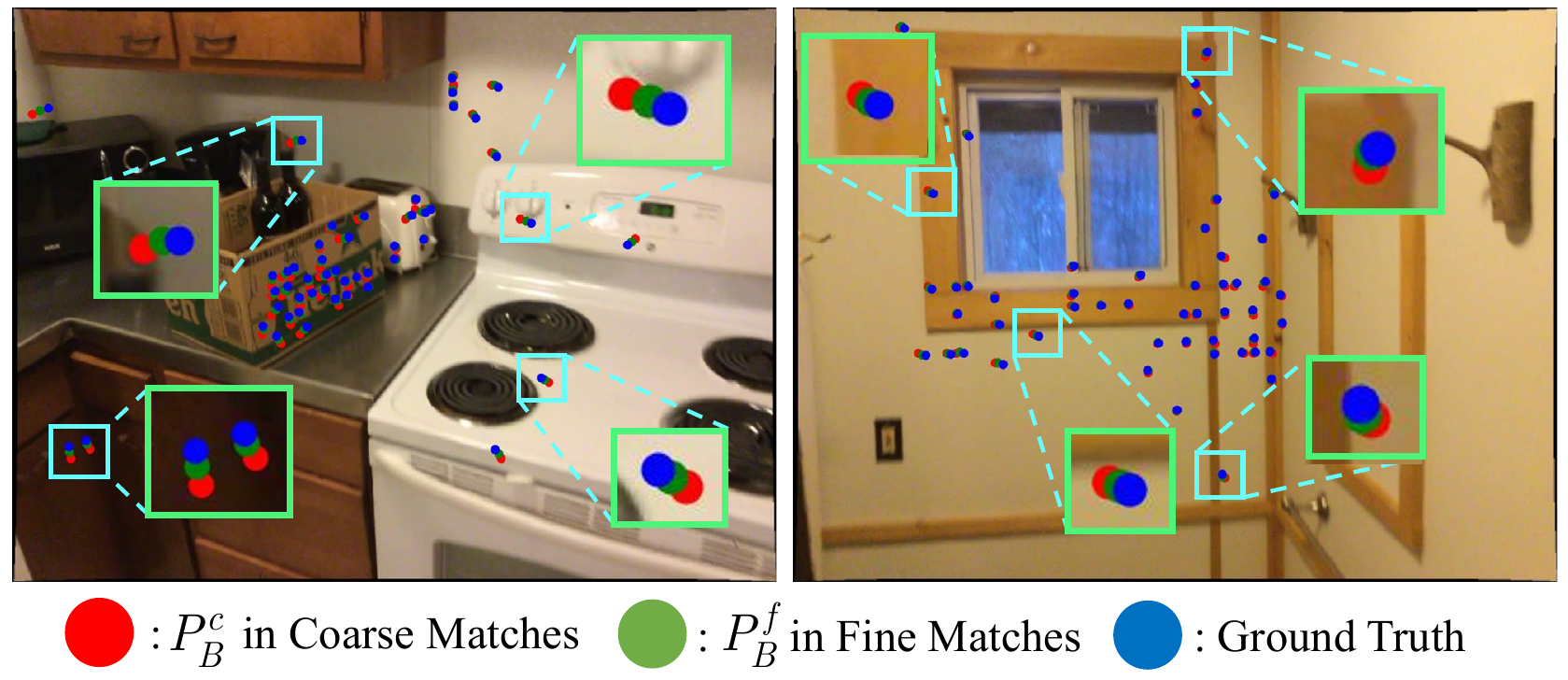}
		\caption{\textbf{Visualization of refinement.} Compared with the coarse matches, the fine matches are closer to the ground-truth.
		}
		\label{fine_adjust}
		\vspace{-10pt}
	\end{figure}

	\subsection{Matches Refinement Block (MRB)}
	To optimize the coarse matches, LoFTR proposes a coarse-to-fine module that only predicts the offset of the coarse matches without appraising whether the predicted matches are reliable.
	Instead, we perceive the match refinement as a combination of classification and regression tasks, and predict the offset with confidence to optimize coarse matches $E^{c}$, hence obtaining precise matches $E^{f}$.
	
	Specifically, for each predicted coarse match, we first locate its position at fine-level features $\bar{F}_{A}, \bar{F}_{B}$ and crop two sets of feature windows $\bar{F}^{w}_{A}, \bar{F}^{w}_{B} \in \mathbb{R}^{T \times \bar{C} \times w \times w}$, where $T$ means the number of predicted matches, $w$ denotes the size of feature windows.
	Then, we reshape $\bar{F}^{w}_{A}, \bar{F}^{w}_{B}$ to $\mathbb{R}^{T \times w^{2} \times \bar{C}}$ to generate sequences, interleave the Transformer Encoder Layer by $L_{3}$ times for feature enhancement, and convert the sequences to feature maps $^{L}\bar{F}^{w}_{A}, ^{L}\bar{F}^{w}_{B} \in \mathbb{R}^{T \times \bar{C} \times w \times w}$.
	Subsequently, we concatenate the feature maps along the channel dimension and utilize six convolutional layers and a maxpooling layer to predict offset $\theta \in \mathbb{R}^{T \times 2}$ and confidence $c \in \mathbb{R}^{T \times 1}$ of the coarse matches.
	Notably, we utilize sigmoid activation function to handle confidence $\theta$ for normalization.
	\begin{equation}
		\begin{aligned}
			F_{mid} = C_{1}(&C_{1}(MP(C_{1}(C_{1}([^{L}\tilde{F}^{w}_{A}||^{L}\tilde{F}^{w}_{B}]))))) \\
			\theta = C_{1}&(F_{mid}), \ \ c = Sig(C_{1}(F_{mid})),
		\end{aligned}
	\end{equation}
	where $C_{1}(\cdot)$ means $1\times1$ convolutional operation;
	$MP(\cdot)$ means maxpooling operation;
	$Sig(\cdot)$ means sigmoid activation function.
	
	Given the coarse matches $E^{c}$ and the predicted offset $\theta$, we define the fine matches $E^{f} = \{ (P_{A}^{f}, P_{B}^{f}) \}$ as:
	\begin{equation}
		E^{f} = \{(P_{A}^{c}(i), P_{B}^{c}(i) + \theta(i)) \ | \ i\in\{1,2,...,T\} \}
	\end{equation}
	
	\begin{figure}
		\centering
		\includegraphics[width=0.999\hsize]{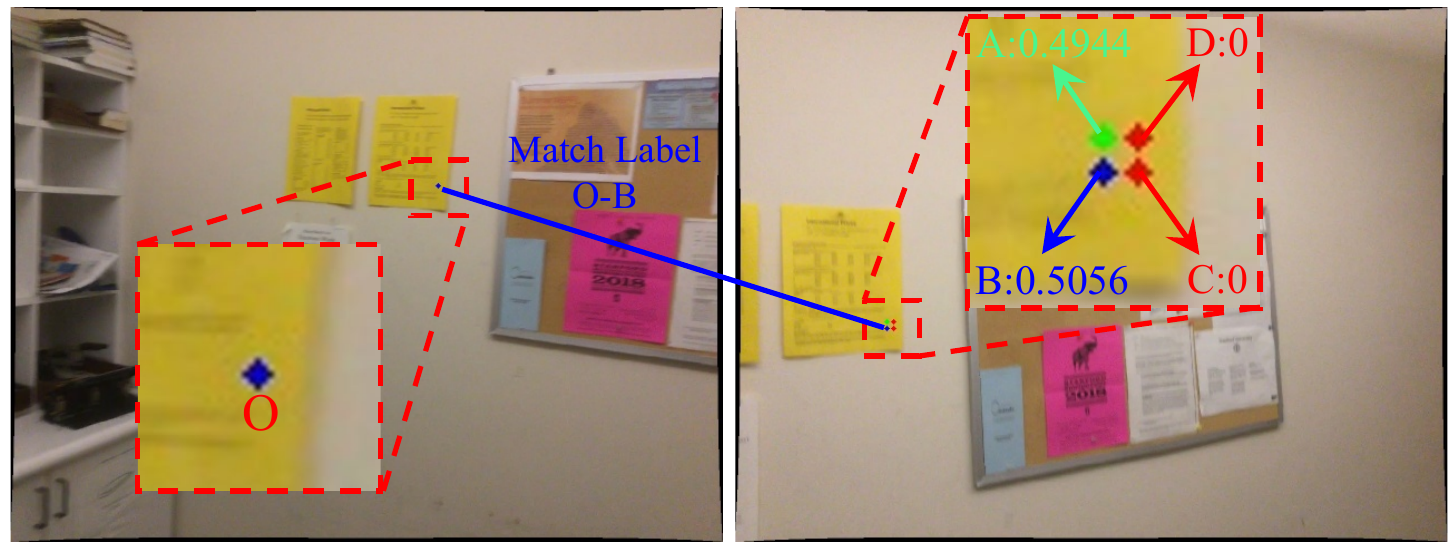}
		\caption{\textbf{The illustration of MLWS.} The match labels $O-A$ and $O-B$ are appended with different label confidence to alleviate the influence of the confused ground-truth match label $O-B$.
		}
		\label{MLWS_fig}
		\vspace{-10pt}
	\end{figure}
	\subsection{Match Labels Weight Strategy (MLWS)}
	Owing to the measurement noise coming from the data, such as the inaccurate depth values, humans prefer to utilize probability to judge whether the match labels are reliable or not, hence relieving the influence of the unreliable matches labels.
	However, existing detector-free methods only utilize $0$ or $1$ to appraise the match labels, which encumbers the learning process.
	To tackle this issue, we propose a Match Labels Weight Strategy (MLWS) that predicts label confidence $G^{lc} \in \mathbb{R}^{N \times N}$ to weight loss function.
		
	
	Concretely, following LoFTR, we first extract the keypoints that satisfy mutual nearest neighbor (MNN) criteria as ground-truth match labels $G^{gt}$.
	Then, as shown in \cref{MLWS_fig}, take the $i$-th extracted keypoint $O$ in the first image as example, we wrap the keypoint $O$ to the second image to obtain projection point, which locates in a grid that utilizes four keypoints (i.e., $A, B, C, D$) as corners.
	Then, we calculate the Euclidean distance between the projection point and four corners of the grids, deriving the distance $d_{1}, d_{2}, d_{3},$ and $d_{4}$, in which the maximum is defined as $d_{m}$.
	Subsequently, we define the scores for the two corners (i.e., $A, B$) with less distance as $d_{m}-d_{1}, d_{m}-d_{2}$, which are handled by softmax to calculate the match probability $P_{1}, P_{2}$ between the $i$-th keypoint $O$ in the first image and the two selected keypoints $A, B$ in the second image:
	\begin{equation}
		\begin{aligned}
			P_{1}, P_{2} = Softma&x((d_{m}-d_{1})/\kappa, \ (d_{m}-d_{2})/ \kappa), \\
		\end{aligned}
	\end{equation}
	where $\kappa$ means temperature coefficient.
	
	Ultimately, we define the corresponding match probability in the $i$-th row of the label confidences $G^{lc}_{A}$ as:
	\begin{equation}
		G^{lc}_{A}(i,ind_{1}) = P_{1}, \ \ G^{lc}_{B}(i,ind_{2}) = P_{2},
	\end{equation}
	where $G^{lc}_{A}$ means the label confidences for the first image;
	$ind_{1}, ind_{2}$ denote the indexes of the selected two keypoints $A, B$ in the second image.
	Other values in the $i$-th row of $G^{lc}_{A}$ are set to $0$.
	
	We perform the same procedures for the keypoints in the second image, obtaining $G^{lc}_{B}$. 
	Finally, we define the label confidences $G^{lc}$ as:
	\begin{equation}
		G^{lc} = (G^{lc}_{A} + G^{lc}_{B}) \ / \ 2
	\end{equation}
	
	The label confidences $G^{lc}$ are utilized to weight loss function, as illustrated in \cref{4800_loss}.
	Besides, as shown in \cref{MLWS_fig}, $G^{lc}$ effectively alleviates the influence of the confused ground-truth match label $O-B$ through attaching different confidences to match labels $O-A$ and $O-B$.

	\subsection{Loss}
	OAMatcher captures the keypoints in overlapping areas to extract coarse matches, which are optimized by the predicted offset $\theta$ to generate final fine matches.
	Therefore, the total loss $L^{t}$ of OAMatcher is comprised of four component:
	(i) The entire image match loss $L^{e}$ for soft assigenment matrix $G$,
	(ii) The overlapping areas match loss $L^{a}$ for co-visible soft assigenment matrix $G^{oa}$,
	(iii) The offset loss $L^{o}$ for $\theta$,
	(iv) The confidence loss $L^{c}$ for $c$.
	\begin{equation}
		L^{t} = \alpha_{1}L^{e} + \alpha_{2}L^{a} + \alpha_{3}L^{o} + \alpha_{4}L^{c}
	\end{equation}
	where $\alpha_{1}, \alpha_{2}, \alpha_{3}, \alpha_{4}$ denote weighting coefficients used to reconcile different losses.
	
	\textbf{Entire Image Match Loss $L^{e}$.}
	Since soft assigenment matrix $G$ directly determines the precision of the predicted co-visible mask $CM_{A}, CM_{B}$, we utilize the ground-truth match labels $G^{gt}$ to supervise $G$ with label confidences $G^{lc}$ as weighting coefficients:
	
	\begin{equation}
		\begin{aligned}
			L^{m}&= -[\frac{1}{|I|}\sum_{(i,j)\in I}G^{lc}(i,j)logG(i,j) + \\
			&\frac{1} {|\tilde{I}|}\sum_{(i,j)\in \tilde{I}} (1-G^{lc}(i,j))log(1-G(i,j))]
		\end{aligned}
		\label{4800_loss}
	\end{equation}
	where $I = \{ (i,j) \ | \ G^{gt}(i,j) = 1 \}$; $\tilde{I} = \{ (i,j) \ | \ G^{gt}(i,j) = 0 \}$;
	$|\cdot|$ means the length of the sequences.
	
	\textbf{Overlapping Areas Match Loss $L^{a}$.}
	In accordance with the index of the co-visible keypoints $P_{A}^{oa}, P_{B}^{oa}$ in the original keypoints $P_{A}, P_{B}$, we obtain the ground-truth match labels and label confidences $G^{gt}_{oa}, G^{lc}_{oa} \in \mathbb{R}^{K \times K}$ for co-visible keypoints.
	Then, similar to \cref{4800_loss}, we utilize $G^{gt}_{oa}, G^{lc}_{oa}$ to supervise $G^{oa}$, deriving the overlapping areas match loss $L^{a}$.
	
	\textbf{Offset Loss $L^{o}$.}
	We utilize offset loss to ensure that the network can effectively optimize the coarse matches.
	Specifically, for the predicted match $(P_{A}^{f}, P_{B}^{f})$, we warp the keypoints $P_{A}^{f}$ in the first image to the second image according to the depth value and the known camera pose, obtaining the ground-truth keypoints $P_{B}^{gt} \in \mathbb{R}^{K \times 2}$.
	Then, we define the ground-truth offset $\theta^{gt}$ as:
	\begin{equation}
		\theta^{gt} = P_{B}^{gt} - P_{B}^{f}
	\end{equation}
	
	Given the predict offset $\theta$ and the ground-truth offset $\theta^{gt}$, we define the offset loss $L^{o}$ as:
	\begin{equation}
		\begin{split}
			L^{o} = \frac{1}{T}\sum_{i=1}^{T}\Vert \theta^{gt}(i)-\theta(i) \Vert_2^2,
		\end{split}
	\end{equation}
	where $T$ is the number of predicted matches.
	Notably, we discard the predicted matches with $\theta^{gt}$ larger than a predefined threshold $\eta$ to prevent the network from optimizing in a wrong direction.
	
	\textbf{Confidence Loss $L^{c}$.}
	We perceive the predicted matches with $\theta^{gt}$ less than $\eta$ as positive and define the ground-truth confidence labels as $1$.
	For other matches with $\theta^{gt}$ larger than $\eta$, we view them as negetive and define the ground-truth confidence labels as $0$.
	According to the ground-truth confidence labels $c^{gt}$ and the predicted confidence $c$, we define the confidence loss $L^{c}$ as binary cross entropy loss:
	\begin{equation}
		L^{c} = -\frac{1}{T}\sum_{i=1}^{T}\left[c^{gt}(i)log \ c(i) + (1-c^{gt}(i))log \ (1-c(i))\right]
	\end{equation} 
	
	\section{EXPERIMENTS} \label{S4}
	\subsection{Implementation Details}
	\textbf{Architecture Details.} 
	The dimensions of the fine-level and coarse-level feature maps are $\bar{C} = 128, \tilde{C} = 256$.
	The scale rate $\gamma$ in Feed-forward Networt is set to $4$.
	We interleave the Transformer Encoder Layer by $L_{1} = 2, \ L_{2} = 2, \ L_{3} = 2$ for features enhancement.
	The kernel size of morphological close operation is set to $k = 10$ for both indoor and outdoor scenes.
	Following SuperGlue, we set the confidence threshold $\rho = 0.2$ to obtain coarse matches.
	Besides, we choose $w=5$ to crop local windows in fine-level feature maps for match refinement.
	When making label confidences, we set $\kappa = 0.01$.
	To reconcile the proposed four losses, we set $\alpha_{1}, \alpha_{2} = 1$, $\alpha_{3}, \alpha_{4} = 0.2$.
	The threshold $\eta$ in offset loss and confidence loss is set to $8$.
	
	\textbf{Training Scheme for Scannet.}
	We utilize $32$ Tesla V100 GPUs to train OAMatcher on Scannet \cite{dai2017scannet} for indoor local feature matching with a batch size of $32$.
	We employ the AdamW solver \cite{loshchilov2017decoupled} for optimization with a weight decay of $0.1$. 
	The initial learning rate is set to $6 \times 10^{-3}$, with a linear warm up for $4800$ iterations.
	Then, the learning rate drop by $0.5$ every $3$ epochs.
	We use gradient clipping that is set to $0.5$ to avoid exploding gradients.
	The total epochs are set to $30$.
	
	\textbf{Training Scheme for MegaDepth.}
	We utilize $32$ Tesla V100 GPUs to train OAMatcher on MegaDepth \cite{li2018megadepth} for outdoor local feature matching with a batch size of $32$.
	We employ the AdamW solver for optimization with a weight decay of $0.1$.
	The initial learning rate is set to $8 \times 10^{-3}$, with a linear learning rate warm-up in $3$ epochs from $8 \times 10^{-4}$ to the initial learning rate.
	Then, the learning rate drop by $0.5$ every $4$ epochs.
	The total epochs are set to $30$.

	\subsection{Homography Estimation}
	To verify the capability of OAMatcher to effectively estimate the geometry relations of image pairs, we conduct a homography estimation experiment.
	
	\textbf{Dataset.}
	We conduct the homography estimation experiment on HPatches dataset \cite{balntas2017hpatches}, which consists of $52$ sequences with significant illumination changes and $56$ sequences that exhibit extreme viewpoint changes. 
	Each sequence is comprised of a ground-truth homography matrix and $6$ images with progressively larger appearance variants.
	
	\textbf{Evaluation Protocol.}
	Following the corner correctness metric (CCM) employed in \cite{zhou2021patch2pix}, we report the proportion of image pairs with average corner errors $\varepsilon$ less than 1/3/5 pixels.
	Specificially, we utilize \textit{findHomography} function to calculate the homography matrix $\hat{H}$ of image pairs.
	Then, according to the predicted homography matrix $\hat{H}$ and ground-truth homography matrix $H$, we warp the four corners of the first image to the second image and calculate the average corner error as corner error.
	Ultimately, we count the proportion of image pairs with average corner errors less than predefined thresholds.
	
	\begin{table}[]\huge
		\centering
		\renewcommand\arraystretch{1.40}
		\caption{\textbf{Homography estimation evaluation} on HPatches dataset.}
		\resizebox{0.48\textwidth}{!}{
			\begin{tabular}{llccc}
				\toprule[2pt]
				\multicolumn{1}{c}{\multirow{2}{*}{\begin{tabular}[c]{@{}c@{}}Local\\ features\end{tabular}}} & \multicolumn{1}{c}{\multirow{2}{*}{Matcher}} & \multicolumn{1}{c}{Overall} & Illumination   & Viewpoint      \\ \cline{3-5} 
				\multicolumn{1}{c}{}                                                                          & \multicolumn{1}{c}{}                         & \multicolumn{3}{c}{CCM ($\varepsilon$\textless 1/3/5 pixels)}              \\ \hline
				\multicolumn{5}{c}{\multirow{1}{*}{Detector-based Methods}}
				\\ \hline
				D2-Net \cite{Dusmanu_2019_CVPR}                                                                                        & NN                                           & 0.38/0.71/0.82              & 0.66/\textbf{0.95}/\textbf{0.98} & 0.12/0.49/0.67 \\
				\multirow{4}{*}{SuperPoint \cite{DeTone_2018_CVPR_Workshops} }
				& NN                                           &0.46/0.78/0.85              & 0.57/0.92/0.97 & 0.35/0.65/0.74 \\
				& SuperGlue \cite{sarlin2020superglue}                                   & 0.51/0.82/0.89              & 0.60/0.92/\textbf{0.98} & 0.42/0.71/0.81 \\
				& SGMNet \cite{chen2021learning}                                         & 0.52/\textbf{0.85}/\textbf{0.91}              & 0.59/0.94/\textbf{0.98} & \textbf{0.46}/0.74/\textbf{0.84} \\ 
				& ClusterGNN \cite{shi2022clustergnn}                                         & 0.52/0.84/0.90              & 0.61/0.93/\textbf{0.98} & 0.44/0.74/0.81 \\ 
				\hline
				\multicolumn{5}{c}{\multirow{1}{*}{Detector-free Methods}}
				\\ \hline
				\multirow{5}{*}{------}                                               
				& SparseNCNet \cite{rocco2020efficient}                                 & 0.36/0.65/0.76              & 0.62/0.92/0.97 & 0.13/0.40/0.58 \\
				& Patch2Pix \cite{zhou2021patch2pix}                                   & 0.50/0.79/0.87              & 0.71/\textbf{0.95}/\textbf{0.98} & 0.30/0.64/0.76 \\
				& LoFTR \cite{sun2021loftr}                                         & \textbf{0.55}/0.81/0.86              & 0.74/\textbf{0.95}/\textbf{0.98} & 0.38/0.69/0.76 \\
				& MatchFormer \cite{wang2022matchformer}                                 & \textbf{0.55}/0.81/0.87              & \textbf{0.75}/\textbf{0.95}/\textbf{0.98} & 0.37/0.68/0.78 \\
				& OAMatcher                                        & 0.54/\textbf{0.85}/\textbf{0.91}           & 0.67/\textbf{0.95}/\textbf{0.98}  & 0.42/\textbf{0.75}/\textbf{0.84} \\ \bottomrule[2pt]
		\end{tabular}}
		\label{homo_experiment}
	\end{table}

	\begin{figure}
		\centering
		\includegraphics[width=0.99\hsize]{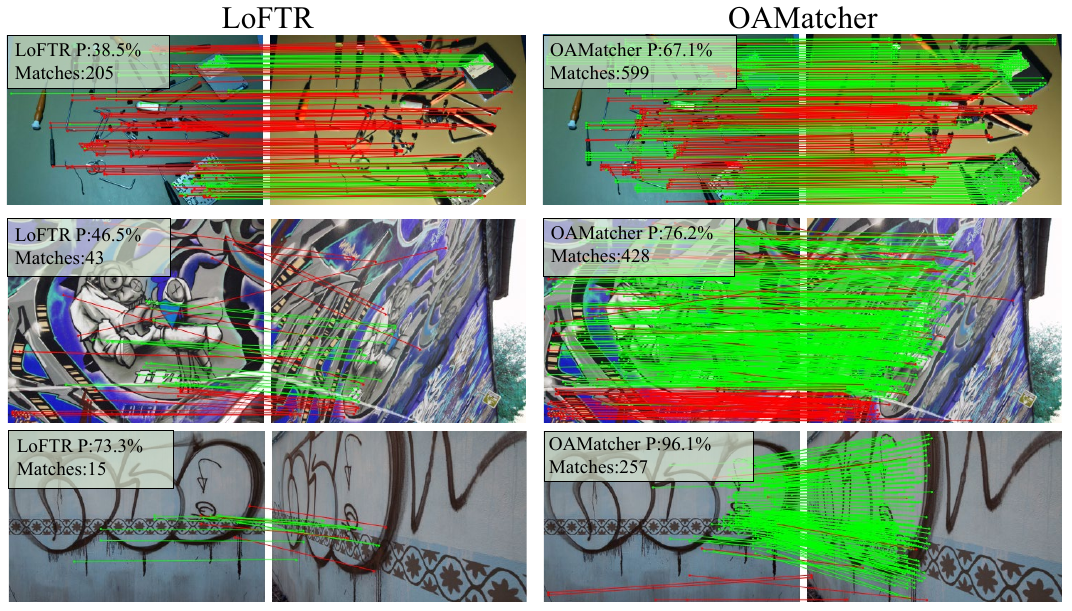}
		\caption{\textbf{Matches predicted by LoFTR and OAMatcher.} OAMatcher realizes superior performance when handling image pairs with extreme appearance changes.
		}
		\label{HPatcher_line}
		\vspace{-4pt}
	\end{figure}
	\textbf{Results.}
	As shown in \cref{homo_experiment}, compared with all detector-based and detector-free methods, OAMatcher achieves outstanding performance.
	Specifically, OAMatcher surpasses the baseline LoFTR with the improvement of $(4\%, 5\%)$ under the threshold of $3,5$ pixels.
	Besides, we can observe that the detector-based methods realize superior performance when handling image pairs with extreme viewpoint variations, while the detector-free methods are more robust to illumination changes.
	However, as detector-free methods, OAMatcher realizes impressive accuracy when handling image pairs with both extreme illumination and viewpoint changes when thresholds are set to $3,5$ pixels.
	Notably, OAMatcher exhibits superior performance under large viewpoint changes among detector-free methods.
	Concretely, OAMatcher outperforms LoFTR and MatchFormer by $(4\%, 6\%, 8\%)$ and $(5\%, 7\%, 6\%)$ under viewpoint changes.
	To further validate the robustness of OAMatcher, we select three image pairs from HPatches and visualize the predicted matches, as shown in \cref{HPatcher_line}.
	Compared with the baseline LoFTR, OAMatcher generate more precise and denser matches.

	\subsection{Indoor Pose Estimation}
	Indoor pose estimation remains a challenging task owing to the sparse texture, motion blur and significant viewpoint shifts.
	Therefore, we conduct an indoor pose estimation experiment to verify the effectiveness of OAMatcher.
	
	\textbf{Dataset.}
	We conduct the indoor pose estimation experiment on Scannet, a large-scale dataset comprising $1513$ sequences with RGB-D images and corresponding ground-truth poses.
	Following \cite{sarlin2020superglue}, we select $230$ million image pairs with the size of $640 \times 480$ as the training set and $1500$ image pairs as the testing set.
	
	\textbf{Evaluation Protocol.}
	Following \cite{sun2021loftr, sarlin2020superglue}, we report the area under the cumulative curve (AUC) of the pose errors at thresholds $(5^{\circ}, 10^{\circ}, 20^{\circ})$.
	Specifically, we leverage OPENCV to calculate the essential matrix and relative pose of image pairs.
	Subsequently, we count the pose errors (i.e., the maximum angular errors of rotation and translation) of all image pairs and plot the cumulative error distribution curve, whose area at three thresholds $(5^{\circ}, 10^{\circ}, 20^{\circ})$ are computed as AUC@$(5^{\circ}, 10^{\circ}, 20^{\circ})$, respectively.
	
	\begin{table}[]
		\centering
		\renewcommand\arraystretch{1.2}
		\caption{\textbf{Indoor pose estimation evaluation} on Scannet dataset. The AUC@(5$^\circ$, 10$^\circ$, 20$^\circ$) is reported. }
		\label{table:1a}
		\resizebox{0.48\textwidth}{!}{
			\begin{tabular}{clccc}
				\toprule[0.3mm]
				\multicolumn{1}{c}{\multirow{2}{*}{\begin{tabular}[c]{@{}c@{}}Local\\ features\end{tabular}}} & \multicolumn{1}{c}{\multirow{2}{*}{Matcher}} & \multicolumn{3}{c}{Pose estimation AUC}          \\ \cline{3-5} 
				\multicolumn{1}{c}{}                                                                          & \multicolumn{1}{c}{}                         & @5$^{\circ}$          & @10$^{\circ}$         & @20$^{\circ}$         \\ \hline
				\multicolumn{5}{c}{\multirow{1}{*}{Detector-based Methods}}
				\\ \hline
				D2-Net \cite{Dusmanu_2019_CVPR}                                                                                       & NN                                  & 5.25           & 14.53          & 27.96          \\
				\multirow{6}{*}{SuperPoint \cite{DeTone_2018_CVPR_Workshops} }                                                                   & NN                                  & 9.43           & 21.53          & 36.40          \\
				& OANet \cite{zhang2019learning}                                  & 16.16          & 33.81          & 51.84          \\
				& SuperGlue \cite{sarlin2020superglue}                                  & 16.16          & 33.81          & 51.84          \\
				& SGMNet \cite{chen2021learning}                                       & 15.40          & 32.06          & 48.32          \\
				& DenseGAP \cite{kuang2021densegap}                                     & 17.01          & 36.07          & 55.66          \\ 
				& HTMatch \cite{cai2022htmatch}                                     & 15.11          & 31.42          & 48.23          \\\hline
				\multicolumn{5}{c}{\multirow{1}{*}{Detector-free Methods}}
				\\ \hline
				\multirow{5}{*}{------}  
				& DRCNet \cite{sun2021loftr}                  & 7.69  & 17.93  & 30.49          \\
				& LoFTR \cite{sun2021loftr}                  & 22.06 & 40.80          & 57.62          \\
				& QuadTree \cite{tang2022quadtree}                  & 24.90 & 44.70          & 61.80          \\
				& MatchFormer \cite{wang2022matchformer}                                        & 24.31          & 43.90 & 61.41 \\
				& OAMatcher                                      & \textbf{26.06}          & \textbf{45.34} & \textbf{62.08} \\
				\bottomrule[0.3mm]
		\end{tabular}}
		\label{Scannet}
	\end{table}

	\textbf{Results.}
	As shown in \cref{Scannet}, the detector-free methods achieve superior performance than detector-based approaches since the detector struggles to extract repeatable keypoints when handling image pairs with significant appearance variants. 
	Compared with other state-of-the-arts, OAMatcher achieves impressive performance.
	Specifically, OAMatcher surpasses the cutting-edge detector-based method DenseGAP by $(9.05\%, 9.27\%, 6.42\%)$ in terms of AUC@$(5^\circ, 10^\circ, 20^\circ)$.
	Besides, OAMatcher achieves superior matching performance than LoFTR, QuadTree and MatchFormer with the improvement of $(4.00\%$,$4.54\%$,$4.46\%)$, $(1.16\%$,$0.64\%$,$0.28\%)$, $(1.75\%$,$1.44\%$,$0.67\%)$.
	Moreover, compared with QuadTree, OAMatcher only consumes $93.32\%$ GFLOPs with $38.16\%$ inference speed boost, as illustrated in \cref{Analysis_Section}.

	\begin{table}[] \small
		\centering
		\renewcommand\arraystretch{1.2}
		\caption{\textbf{Outdoor pose estimation evaluation} on MegaDepth dataset. 
			The AUC@(5$^\circ$, 10$^\circ$, 20$^\circ$) is reported.}
		\label{table:2a}
		\resizebox{0.46\textwidth}{!}{
			\begin{tabular}{clccc}
				\toprule[1pt]
				\multicolumn{1}{c}{\multirow{2}{*}{\begin{tabular}[c]{@{}c@{}}Local\\ features\end{tabular}}} & \multicolumn{1}{c}{\multirow{2}{*}{Matcher}} & \multicolumn{3}{c}{Pose estimation AUC}          \\ \cline{3-5} 
				\multicolumn{1}{c}{}                                                                          & \multicolumn{1}{c}{}                         & @5$^{\circ}$          & @10$^{\circ}$         & @20$^{\circ}$         \\ \hline
				\multicolumn{5}{c}{\multirow{1}{*}{Detector-based Methods}}
				\\ \hline
				\multirow{3}{*}{SuperPoint\cite{DeTone_2018_CVPR_Workshops} }                                                                   & SuperGlue \cite{sarlin2020superglue}                                       & 42.18          & 61.16          & 75.96          \\
				& DenseGAP \cite{kuang2021densegap}                                     & 41.17          & 56.87          & 70.22          \\ 
				& ClusterGNN \cite{shi2022clustergnn}                                     & 44.19          & 58.54          & 70.33          \\ \hline
				\multicolumn{5}{c}{\multirow{1}{*}{Detector-free Methods}}
				\\ \hline
				\multirow{7}{*}{------}   
				& DRCNet \cite{sun2021loftr}                  & 27.01  & 42.96  & 58.31          \\ 
				& LoFTR \cite{sun2021loftr}                                      & 52.80 & 69.19 & 81.18          \\
				& QuadTree \cite{tang2022quadtree}                                       & 54.60          & 70.50          & 82.20
				\\ 
				& TopicFM \cite{truong2022topicfm}                                       & 54.10          & 70.10          & 81.60
				\\ 
				& MatchFormer \cite{wang2022matchformer}                                       & 52.91          & 69.74          & 82.00
				\\ 
				& ASpanFormer \cite{chen2022aspanformer}                                         & 55.30          & 71.50 & 83.10 \\
				& OAMatcher                                         & \textbf{56.56}          & \textbf{72.34} & \textbf{83.61} \\
				\bottomrule[1pt]
		\end{tabular}}
		\label{MegaDepth}
	\end{table}

	\subsection{Outdoor Pose Estimation}
	Outdoor images present more extreme challenges, such as occlusion, illumination and viewpoint changes.
	To verify the efficacy of OAMatcher in overcoming these issues, we conduct an outdoor pose estimation experiment.
	
	\textbf{Dataset.}
	We utilize MegaDepth \cite{li2018megadepth} to conduct the outdoor pose estimation experiment.
	MegaDepth consists of $196$ 3D reconstructions with 1M internet images, whose camera poses and depth maps are created by COLMAP \cite{schonberger2016structure}.
	Following \cite{sun2021loftr, tyszkiewicz2020disk}, we select 100 image pairs each scene for training and 1500 image pairs for testing. 
	We resize the longer dimension of images to $840$ when training and testing.
	
	\textbf{Evaluation Protocol.}
	Following \cite{sun2021loftr}, we use the same evaluation metrics AUC@(5$^\circ$, 10$^\circ$, 20$^\circ$) as the indoor pose estimation task.
	
	\textbf{Results.}
	As shown in \cref{MegaDepth}, we can also observe that the detector-free methods consistently surpass the detector-based approaches by a significant margin.
	Among all of the state-of-the-arts, OAMatcher consistently achieves impressive performance under all error thresholds.
	Specifically, OAMatcher outperforms the cutting-edge detector-based method SuperGlue and ClusterGNN by $(14.38\%,$$11.18\%,$$7.65\%)$, $(12.37\%,$$13.80\%,$$13.28\%)$.
	Benefiting from the global receptive field of Transformer, OAMatcher significantly surpasses the CNN-based method DRCNet.
	Besides, compared with the baseline LoFTR, OAMatcher achieves superior performance with the improvement of $(3.76\%,$$3.15\%,$$2.43\%)$.
	Moreover, OAMatcher outperforms the state-of-the-art detector-free methods QuadTree, TopicFM, MatchFormer, and ASpanFormer by $(1.96\%,$$1.84\%,$$1.41\%)$, $(2.46\%,$ $2.24\%,$$2.01\%)$, $(3.65\%,$$2.60\%,$$1.61\%)$, $(1.26\%,$$0.84\%,$$0.51\%)$, demonstrating the superior matching capability of OAMatcher in outdoor pose estimation task.

	\subsection{OAMatcher Analysis}
	\begin{table}[] \huge
		\centering
		\renewcommand\arraystretch{1.2}
		\caption{\textbf{Efficiency analysis.} Several approaches are compared in terms of parameters (MB), GFLOPS and runtime (s).}
		\resizebox{0.35\textwidth}{!}{
			\begin{tabular}{l|ccc}
				\toprule[2.5pt]
				\multicolumn{1}{c|}{Methods} & Params & GFLOPs & Runtime \\ \midrule
				LoFTR \cite{sun2021loftr}                        & \textbf{11.06}    & \textbf{328.67}  & \textbf{0.079}       \\
				QuadTree \cite{tang2022quadtree}               & 13.21  & 382.01     & 0.152       \\
				MatchFormer \cite{wang2022matchformer}                  & 22.37   & 396.95    & 0.357       \\
				AspanFormer \cite{chen2022aspanformer}                 & 15.05    & 391.38   & 0.141       \\
				OAMatcher                  & 15.29     & 356.48      & 0.094           \\
				\bottomrule[2.5pt]
			\end{tabular}
		}
		\label{time_flops}
	\end{table}
	
	\begin{figure*}
		\centering
		\includegraphics[width=1.00\hsize]{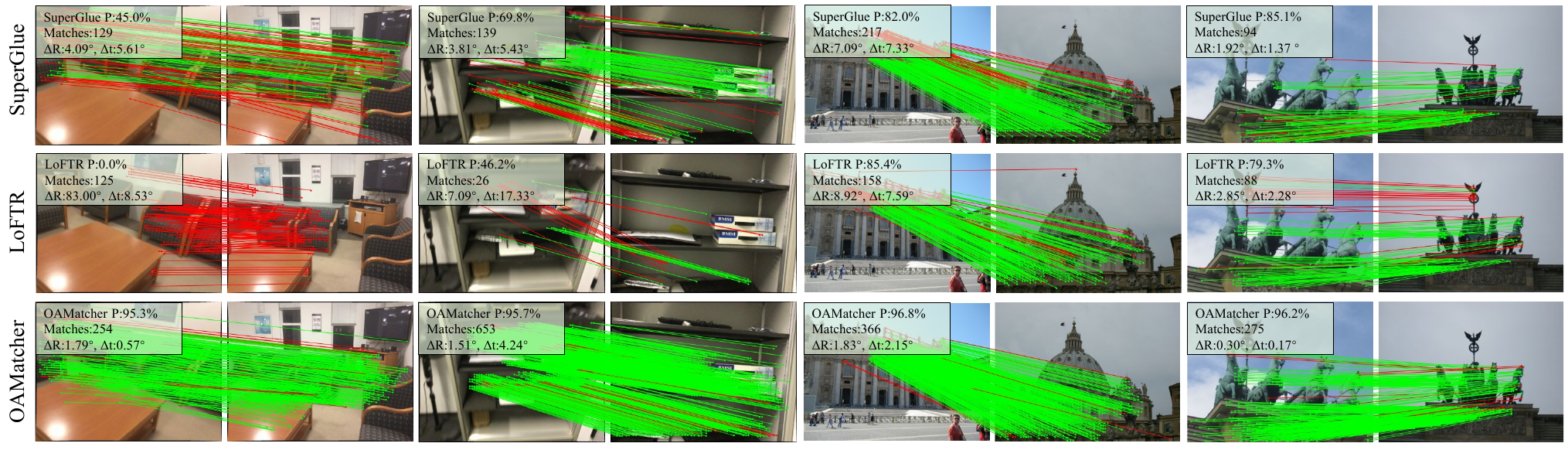}
		\caption{\textbf{A qualitative comparison between SuperGlue, LoFTR, and OAMatcher.} The correspondences with reprojection errors larger than $15$ pixels are colored by red. 
		}
		\label{matches}
	\end{figure*}
	
	\begin{figure*}
		\centering
		\includegraphics[width=0.99\hsize]{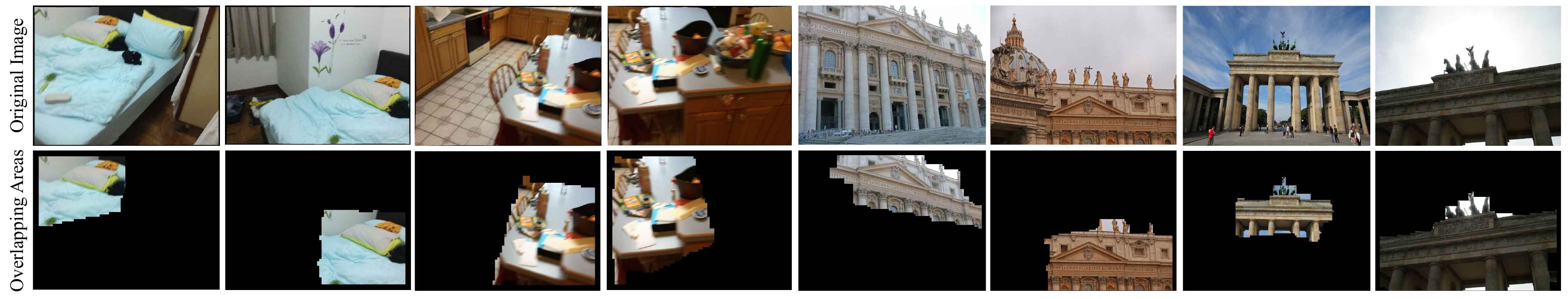}
		\caption{Visualization of the \textbf{predicted overlapping areas} of image pairs with extreme appearance variants. 
		}
		\label{mask_outdoor}
	\end{figure*}
	\textbf{Efficiency Analysis.} \label{Analysis_Section}
	To analysis the efficiency of OAMatcher, we compare several state-of-the-art detector-free methods in terms of parameters, flops and inference speed.
	We resize the input images to $640 \times 480$ and conduct all experiments on a single Tesla V100.
	When counting runtime, we count the time of processing all test images in Scannet and report the average time to eliminate occasionality.
	As shown in \cref{time_flops}, LoFTR possesses less models parameters and faster inference speed, while the pose estimation accurate is inferior, as demonstrated in \cref{Scannet}, \cref{MegaDepth}.
	Besides, compared with the cutting-edge approaches QuadTree, MatchFormer and AspanFormer, OAMatcher consumes $(93.32\%, 89.80\%, 91.08\%)$ GFLOPs with $(38.16\%, 73.67\%, 33.33\%)$ inference speed boost.
	
	\textbf{Matching Results Visualization.}
	We conduct a qualitative experiment through visualizing the matching results of SuperGlue, LoFTR, and OAMatcher to exhibit the superior performance of our method.
	As illustrated in \cref{matches}, we select four indoor and outdoor image pairs with extreme appearance variants to perform feature matching, with the results showing that OAMatcher produces more accurate and dense matches, especifically when handling image pairs with large viewpoint changes (i.e., the first image pair).
	
	\begin{figure}
		\centering
		\includegraphics[width=0.99\hsize]{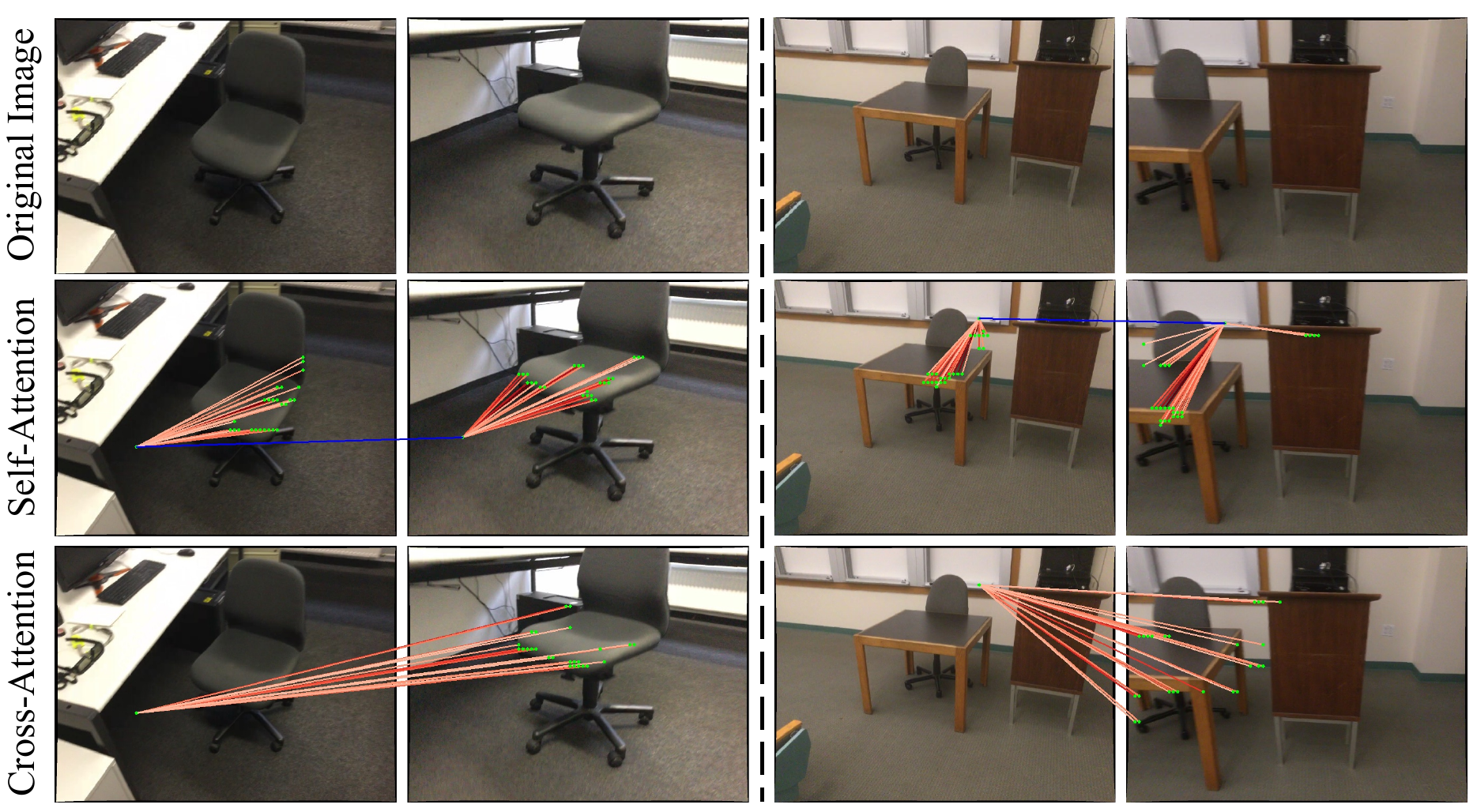}
		\caption{Visualization of the \textbf{self and cross attention weights,} which are shown as rays.
		The keypoints can flexibly aggragate the global context information to locate themselves in the images.}
		\label{attention_line}
	\end{figure}
	\textbf{Overlapping Areas Visualization.}
	To further validate the effectiveness of OAMatcher to generate precise overlapping areas in extreme conditons, we visualize the predicted co-visible regions of four indoor and outdoor image pairs with large illumination and viewpoint changes.
	As shown in \cref{mask_outdoor}, we can observe that OAMatcher captures satisfactory overlapping areas of image pairs
	
	\textbf{Self/Cross Attention Analysis.}
	OAMatcher interleaves the Transformer Encoder Layers to aggregate information intra/inter images.
	To investigate which keypoints provide the prominent information, we visualize the self and cross attention weights, as shown in \cref{attention_line}.
	We can observe that OAMatcher primarily aggregates the global context information from the keypoints in boundaries or corners, which contains abundant visual cue.
	This phenomenon is consistent with the humans behavior that we focuses on the corner and the edge to locate a keypoint in the image.
	
	
	\begin{table}[t]
		\centering
		\renewcommand\arraystretch{1.2}
		\caption{\textbf{Ablation study with different variants of OAMatcher} on MegaDepth dataset.}
		\label{struct_ablation}
		\resizebox{0.46\textwidth}{!}{ 
			\begin{tabular}{lccc}
				\toprule[1.25pt]
				\multicolumn{1}{c}{\multirow{2}{*}{Methods}}           & \multicolumn{3}{c}{Pose estimation AUC} \\ \cline{2-4} 
				& @5$^{\circ}$          & @10$^{\circ}$         & @20$^{\circ}$         \\ \hline
				(i) w only Self-Attention    & 53.06        & 69.51        & 81.57       \\
				(ii) w only Cross-Attention    & 54.61         & 70.94          & 82.45         \\
				(iii) w/o Depth-wise Convolution    & 55.41           & 70.76           & 82.48            \\
				(iv) w/o OAPM     & 54.80           & 70.21          & 81.95            \\
				(v) w/o MLWS  & 55.92            & 71.72           & 83.18            \\
				(vi) w/o MRB    & 52.85            & 68.71           & 80.24          \\
				OAMatcher full                          & \textbf{56.56}          & \textbf{72.34}          & \textbf{83.61}           \\ \bottomrule[1.25pt]
			\end{tabular}
			\label{ablation_experi}
			\vspace{-10pt}}
	\end{table}
	\begin{table}[t]
		\centering
		\renewcommand\arraystretch{1.2}
		\caption{\textbf{Ablation study with different Transformer layers number} on MegaDepth dataset.}
		\label{struct_ablation}
		\resizebox{0.36\textwidth}{!}{ 
			\begin{tabular}{lccc}
				\toprule[1.2pt]
				\multicolumn{1}{c}{\multirow{2}{*}{Methods}}           & \multicolumn{3}{c}{Pose estimation AUC} \\ \cline{2-4} 
				& @5$^{\circ}$          & @10$^{\circ}$         & @20$^{\circ}$         \\ \hline
				$L_{1} = 1, \ L_{2} = 3$    & 55.39        & 71.13        & 82.80      \\
				$L_{1} = 2, \ L_{2} = 2$    & \textbf{56.56}          & \textbf{72.34}          & \textbf{83.61}         \\
				$L_{1} = 3, \ L_{2} = 1$    & 55.71         & 71.78          & 83.21           \\ \bottomrule[1.2pt]
			\end{tabular}
			\label{ablation_transformer}
			\vspace{-10pt}}
	\end{table}
	\subsection{Ablation Study}
	\textbf{Effect of the Proposed Modules.}
	To evaluate the effectiveness of the proposed modules, we conduct ablation experiments on MegaDepth dataset using different variants of OAMatcher, with the results shown in \cref{ablation_experi}.
	(i), (ii) Using only self and cross attention results in a significant performance degradation, demonstrating that it is essential to integrating the information intra/inter images.
	(iii) Discarding the depth-wise convolution in the Transformer Encoder Layer spawns a large drop $(1.15\%, 1.58\%, 1.13\%)$ in AUC values, indicating the significance of extracting local feature representation and global context information concurrently. 
	(iv) Discarding the Overlapping Areas Prediction Module (OAPM) and only performing the global context information interactions among all keypoints by $4$ times results in a large drop $(1.76\%, 2.13\%, 1.66\%)$, proving that OAPM promotes effective and clean message propagation.
	(v) Removing the Match Labels Weight Strategy (MLWS) leads to a degraded pose estimation accuracy $(0.64\%, 0.62\%, 0.43\%)$, proving that using label confidences to weight the loss function effectively alleviates the influence of measurement noise coming from the data.
	(vi) Removing the Matches Refinement Block (MRB) results in a much lower accuracy $(3.71\%, 3.63\%, 3.37\%)$,  indicating the effectiveness of refining coarse matches.
	
	\textbf{Effect of the Transformer Encoder Layers Number $L_{1}, L_{2}$.}
	We interleave the Transformer Encoder Layers by $L_{1}, L_{2}$ times in $EITM$ and $OATM$ to perform features enhancement.
	In this part, we conduct an ablation experiment to obtain the optimal enhancement number $L_{1}, L_{2}$.
	As shown in \cref{ablation_transformer}, we can observe that OAMatcher achieves the best performace when $L_{1} =2, L_{2} = 2$.
	
	\begin{table}[t] \large
		\centering
		\renewcommand\arraystretch{1.2}
		\caption{\textbf{Ablation study with different variants of OAPM} on MegaDepth dataset.}
		\label{struct_ablation}
		\resizebox{0.49\textwidth}{!}{ 
			\begin{tabular}{lccc}
				\toprule[1.5pt]
				\multicolumn{1}{c}{\multirow{2}{*}{Methods}}           & \multicolumn{3}{c}{Pose estimation AUC} \\ \cline{2-4} 
				& @5$^{\circ}$          & @10$^{\circ}$         & @20$^{\circ}$         \\ \hline
				(i) Replacing adaptive threshold with $0.2$    & 55.87        & 71.87       & 83.31       \\
				(ii) Replacing Close Operation with Dilation    & 55.59        & 71.20       & 82.79       \\
				(iii) w/o Finding Maximum Contours    & 55.29         & 71.38          & 83.05        \\ OAMatcher full                           & \textbf{56.56}          & \textbf{72.34}          & \textbf{83.61}            \\ \bottomrule[1.5pt]
			\end{tabular}
			\label{ablation_OAPM}
			\vspace{-10pt}}
	\end{table}

	\textbf{Effect of the Modules in OAPM.}
	To evaluate the rationality of the proposed modules in OAPM, as shown in \cref{ablation_OAPM}, we conduct an ablation experiment using different variants of OAPM.
	(i) Replacing the adaptive threshold with a fixed threshold $\lambda = 0.2$ results in a large drop $(0.69\%, 0.47\%, 0.30\%)$ in AUC values.
	(ii) Replacing the morphological close operation with the dilation operation results in a declined accuracy $(0.97\%, 1.14\%, 0.82\%)$ in terms of AUC@$(5^\circ, 10^\circ, 20^\circ)$.
	(iii) Removing the finding maximum contours operation leads to a performance degradation $(1.27\%, 0.96\%, 0.56\%)$, demonstrating the effectiveness of using maximum counter to eliminate the holes and alleviate the interference coming from noise.
	
%
	
	\begin{table}[t] \scriptsize
		\centering
		\caption{\textbf{Ablation study with different scale rate $\gamma$} on MegaDepth dataset.}
		\renewcommand\arraystretch{1.15}
		\resizebox{0.32\textwidth}{!}{%
			\begin{tabular}{cccc}
				\toprule[0.9pt]
				\multicolumn{1}{c}{\multirow{2}{*}{Methods}} & \multicolumn{3}{c}{Pose estimation AUC}                          \\ \cline{2-4} 
				& @5$^{\circ}$             & @10$^{\circ}$            & @20$^{\circ}$            \\ \hline
				$\gamma$=2                                    & 55.94      & 71.93         & 83.48 \\
				$\gamma$=3                                   & 55.57        & 71.60         & 83.01        \\
				$\gamma$=4                                    & \textbf{56.56}          & \textbf{72.34}          & \textbf{83.61}
				\\ \bottomrule[0.9pt]
		\end{tabular}}
		\label{scalerate_table}
	\end{table}
	\textbf{Effect of the Scale Rate $\gamma$ in FFN.}
	In accordance with conventional Transformer, the scale rate $\gamma$ is essential to extract abundant feature representation.
	To obtain the optimal scale rate, we train OAMatcher with different $\gamma$ and conduct the outdoor pose estimation experiment.
	As shown in \cref{scalerate_table}, we can observe that DeepMatcher achieves the best performance when $\gamma = 4$.

	\section{Conclusion} \label{S5}
	In this work, we propose OAMatcher, a novel detector-free method that imitates humans behavior to realize accurate and dense local feature matching.
	Overlapping Areas Prediction Module is proposed to transit the focus regions of network from entire images to co-visible regions, realizing effective and clean context information aggregation.
	Match Labels Weight Strategy is applied to weight loss using the label confidences, hence alleviating the influence of the unreliable match labels.
	Extensive experiments demonstrate that OAMatcher surpasses state-of-the-art approaches on several benchmarks, demonstrating the excellent performance of OAMatcher.
	
	\bibliographystyle{IEEEtran}
	\bibliography{IEEEabrv,bib.bib}

\begin{thebibliography}{10}
\providecommand{\url}[1]{#1}
\csname url@samestyle\endcsname
\providecommand{\newblock}{\relax}
\providecommand{\bibinfo}[2]{#2}
\providecommand{\BIBentrySTDinterwordspacing}{\spaceskip=0pt\relax}
\providecommand{\BIBentryALTinterwordstretchfactor}{4}
\providecommand{\BIBentryALTinterwordspacing}{\spaceskip=\fontdimen2\font plus
\BIBentryALTinterwordstretchfactor\fontdimen3\font minus
  \fontdimen4\font\relax}
\providecommand{\BIBforeignlanguage}[2]{{%
\expandafter\ifx\csname l@#1\endcsname\relax
\typeout{** WARNING: IEEEtran.bst: No hyphenation pattern has been}%
\typeout{** loaded for the language `#1'. Using the pattern for}%
\typeout{** the default language instead.}%
\else
\language=\csname l@#1\endcsname
\fi
#2}}
\providecommand{\BIBdecl}{\relax}
\BIBdecl

\bibitem{mur2017orb}
R.~Mur-Artal and J.~D. Tard{\'o}s, ``Orb-slam2: An open-source slam system for
  monocular, stereo, and rgb-d cameras,'' \emph{IEEE transactions on robotics},
  vol.~33, no.~5, pp. 1255--1262, 2017.

\bibitem{campos2021orb}
C.~Campos, R.~Elvira, J.~J.~G. Rodr{\'\i}guez, J.~M. Montiel, and J.~D.
  Tard{\'o}s, ``Orb-slam3: An accurate open-source library for visual,
  visual--inertial, and multimap slam,'' \emph{IEEE Transactions on Robotics},
  vol.~37, no.~6, pp. 1874--1890, 2021.

\bibitem{qin2018vins}
T.~Qin, P.~Li, and S.~Shen, ``Vins-mono: A robust and versatile monocular
  visual-inertial state estimator,'' \emph{IEEE Transactions on Robotics},
  vol.~34, no.~4, pp. 1004--1020, 2018.

\bibitem{schonberger2016structure}
J.~L. Schonberger and J.-M. Frahm, ``Structure-from-motion revisited,'' in
  \emph{Proceedings of the IEEE conference on computer vision and pattern
  recognition}, 2016, pp. 4104--4113.

\bibitem{cui2022vidsfm}
H.~Cui, D.~Tu, F.~Tang, P.~Xu, H.~Liu, and S.~Shen, ``Vidsfm: Robust and
  accurate structure-from-motion for monocular videos,'' \emph{IEEE
  Transactions on Image Processing}, vol.~31, pp. 2449--2462, 2022.

\bibitem{zhao2022alike}
X.~Zhao, X.~Wu, J.~Miao, W.~Chen, P.~C. Chen, and Z.~Li, ``Alike: Accurate and
  lightweight keypoint detection and descriptor extraction,'' \emph{IEEE
  Transactions on Multimedia}, 2022.

\bibitem{fan2022seeing}
B.~Fan, Y.~Yang, W.~Feng, F.~Wu, J.~Lu, and H.~Liu, ``Seeing through darkness:
  Visual localization at night via weakly supervised learning of domain
  invariant features,'' \emph{IEEE Transactions on Multimedia}, 2022.

\bibitem{lowe2004distinctive}
D.~G. Lowe, ``Distinctive image features from scale-invariant keypoints,''
  \emph{International journal of computer vision}, vol.~60, no.~2, pp. 91--110,
  2004.

\bibitem{rublee2011orb}
E.~Rublee, V.~Rabaud, K.~Konolige, and G.~Bradski, ``Orb: An efficient
  alternative to sift or surf,'' in \emph{2011 International conference on
  computer vision}.\hskip 1em plus 0.5em minus 0.4em\relax Ieee, 2011, pp.
  2564--2571.

\bibitem{DeTone_2018_CVPR_Workshops}
D.~DeTone, T.~Malisiewicz, and A.~Rabinovich, ``Superpoint: Self-supervised
  interest point detection and description,'' in \emph{Proceedings of the IEEE
  Conference on Computer Vision and Pattern Recognition (CVPR) Workshops}, June
  2018.

\bibitem{Dusmanu_2019_CVPR}
M.~Dusmanu, I.~Rocco, T.~Pajdla, M.~Pollefeys, J.~Sivic, A.~Torii, and
  T.~Sattler, ``D2-net: A trainable cnn for joint description and detection of
  local features,'' in \emph{Proceedings of the IEEE/CVF Conference on Computer
  Vision and Pattern Recognition (CVPR)}, June 2019.

\bibitem{revaud2019r2d2}
J.~Revaud, P.~Weinzaepfel, C.~De~Souza, N.~Pion, G.~Csurka, Y.~Cabon, and
  M.~Humenberger, ``R2d2: repeatable and reliable detector and descriptor,'' in
  \emph{NeurIPS}, 2019.

\bibitem{tyszkiewicz2020disk}
M.~J. Tyszkiewicz, P.~Fua, and E.~Trulls, ``Disk: Learning local features with
  policy gradient,'' in \emph{NeurIPS}, 2020.

\bibitem{sarlin2020superglue}
P.-E. Sarlin, D.~DeTone, T.~Malisiewicz, and A.~Rabinovich, ``Superglue:
  Learning feature matching with graph neural networks,'' in \emph{Proceedings
  of the IEEE/CVF conference on computer vision and pattern recognition}, 2020,
  pp. 4938--4947.

\bibitem{chen2021learning}
H.~Chen, Z.~Luo, J.~Zhang, L.~Zhou, X.~Bai, Z.~Hu, C.-L. Tai, and L.~Quan,
  ``Learning to match features with seeded graph matching network,'' in
  \emph{Proceedings of the IEEE/CVF International Conference on Computer
  Vision}, 2021, pp. 6301--6310.

\bibitem{kuang2021densegap}
Z.~Kuang, J.~Li, M.~He, T.~Wang, and Y.~Zhao, ``Densegap: Graph-structured
  dense correspondence learning with anchor points,'' \emph{arXiv preprint
  arXiv:2112.06910}, 2021.

\bibitem{shi2022clustergnn}
Y.~Shi, J.-X. Cai, Y.~Shavit, T.-J. Mu, W.~Feng, and K.~Zhang, ``Clustergnn:
  Cluster-based coarse-to-fine graph neural network for efficient feature
  matching,'' in \emph{Proceedings of the IEEE/CVF Conference on Computer
  Vision and Pattern Recognition}, 2022, pp. 12\,517--12\,526.

\bibitem{detone2018superpoint}
D.~DeTone, T.~Malisiewicz, and A.~Rabinovich, ``Superpoint: Self-supervised
  interest point detection and description,'' in \emph{Proceedings of the IEEE
  conference on computer vision and pattern recognition workshops}, 2018, pp.
  224--236.

\bibitem{luo2020aslfeat}
Z.~Luo, L.~Zhou, X.~Bai, H.~Chen, J.~Zhang, Y.~Yao, S.~Li, T.~Fang, and
  L.~Quan, ``Aslfeat: Learning local features of accurate shape and
  localization,'' in \emph{Proceedings of the IEEE/CVF conference on computer
  vision and pattern recognition}, 2020, pp. 6589--6598.

\bibitem{bay2006surf}
H.~Bay, T.~Tuytelaars, and L.~V. Gool, ``Surf: Speeded up robust features,'' in
  \emph{European conference on computer vision}.\hskip 1em plus 0.5em minus
  0.4em\relax Springer, 2006, pp. 404--417.

\bibitem{tao2002continuous}
Y.~Tao, D.~Papadias, and Q.~Shen, ``Continuous nearest neighbor search,'' in
  \emph{VLDB'02: Proceedings of the 28th International Conference on Very Large
  Databases}.\hskip 1em plus 0.5em minus 0.4em\relax Elsevier, 2002, pp.
  287--298.

\bibitem{bian2017gms}
J.~Bian, W.-Y. Lin, Y.~Matsushita, S.-K. Yeung, T.-D. Nguyen, and M.-M. Cheng,
  ``Gms: Grid-based motion statistics for fast, ultra-robust feature
  correspondence,'' in \emph{Proceedings of the IEEE conference on computer
  vision and pattern recognition}, 2017, pp. 4181--4190.

\bibitem{zhang2019learning}
J.~Zhang, D.~Sun, Z.~Luo, A.~Yao, L.~Zhou, T.~Shen, Y.~Chen, L.~Quan, and
  H.~Liao, ``Learning two-view correspondences and geometry using order-aware
  network,'' in \emph{Proceedings of the IEEE/CVF International Conference on
  Computer Vision}, 2019, pp. 5845--5854.

\bibitem{chen2022csr}
J.~Chen, S.~Chen, X.~Chen, Y.~Dai, and Y.~Yang, ``Csr-net: Learning adaptive
  context structure representation for robust feature correspondence,''
  \emph{IEEE Transactions on Image Processing}, vol.~31, pp. 3197--3210, 2022.

\bibitem{ma2022correspondence}
J.~Ma, Y.~Wang, A.~Fan, G.~Xiao, and R.~Chen, ``Correspondence attention
  transformer: A context-sensitive network for two-view correspondence
  learning,'' \emph{IEEE Transactions on Multimedia}, 2022.

\bibitem{rocco2018neighbourhood}
I.~Rocco, M.~Cimpoi, R.~Arandjelovi{\'c}, A.~Torii, T.~Pajdla, and J.~Sivic,
  ``Neighbourhood consensus networks,'' \emph{Advances in neural information
  processing systems}, vol.~31, 2018.

\bibitem{rocco2020efficient}
I.~Rocco, R.~Arandjelovi{\'c}, and J.~Sivic, ``Efficient neighbourhood
  consensus networks via submanifold sparse convolutions,'' in \emph{European
  Conference on Computer Vision}.\hskip 1em plus 0.5em minus 0.4em\relax
  Springer, 2020, pp. 605--621.

\bibitem{li2020dual}
X.~Li, K.~Han, S.~Li, and V.~Prisacariu, ``Dual-resolution correspondence
  networks,'' \emph{Advances in Neural Information Processing Systems},
  vol.~33, pp. 17\,346--17\,357, 2020.

\bibitem{zhou2021patch2pix}
Q.~Zhou, T.~Sattler, and L.~Leal-Taixe, ``Patch2pix: Epipolar-guided
  pixel-level correspondences,'' in \emph{Proceedings of the IEEE/CVF
  Conference on Computer Vision and Pattern Recognition}, 2021, pp. 4669--4678.

\bibitem{efe2021dfm}
U.~Efe, K.~G. Ince, and A.~Alatan, ``Dfm: A performance baseline for deep
  feature matching,'' in \emph{Proceedings of the IEEE/CVF Conference on
  Computer Vision and Pattern Recognition}, 2021, pp. 4284--4293.

\bibitem{sun2021loftr}
J.~Sun, Z.~Shen, Y.~Wang, H.~Bao, and X.~Zhou, ``Loftr: Detector-free local
  feature matching with transformers,'' in \emph{Proceedings of the IEEE/CVF
  Conference on Computer Vision and Pattern Recognition}, 2021, pp. 8922--8931.

\bibitem{wang2022matchformer}
Q.~Wang, J.~Zhang, K.~Yang, K.~Peng, and R.~Stiefelhagen, ``Matchformer:
  Interleaving attention in transformers for feature matching,'' \emph{arXiv
  preprint arXiv:2203.09645}, 2022.

\bibitem{truong2022topicfm}
K.~Truong~Giang, S.~Song, and S.~Jo, ``Topicfm: Robust and interpretable
  feature matching with topic-assisted,'' \emph{arXiv e-prints}, pp.
  arXiv--2207, 2022.

\bibitem{chen2022aspanformer}
H.~Chen, Z.~Luo, L.~Zhou, Y.~Tian, M.~Zhen, T.~Fang, D.~McKinnon, Y.~Tsin, and
  L.~Quan, ``Aspanformer: Detector-free image matching with adaptive span
  transformer,'' in \emph{European Conference on Computer Vision}.\hskip 1em
  plus 0.5em minus 0.4em\relax Springer, 2022, pp. 20--36.

\bibitem{sun2019guide}
K.~Sun, W.~Tao, and Y.~Qian, ``Guide to match: multi-layer feature matching
  with a hybrid gaussian mixture model,'' \emph{IEEE Transactions on
  Multimedia}, vol.~22, no.~9, pp. 2246--2261, 2019.

\bibitem{vaswani2017attention}
A.~Vaswani, N.~Shazeer, N.~Parmar, J.~Uszkoreit, L.~Jones, A.~N. Gomez,
  {\L}.~Kaiser, and I.~Polosukhin, ``Attention is all you need,'' in
  \emph{Advances in neural information processing systems}, 2017, pp.
  5998--6008.

\bibitem{katharopoulos2020transformers}
A.~Katharopoulos, A.~Vyas, N.~Pappas, and F.~Fleuret, ``Transformers are rnns:
  Fast autoregressive transformers with linear attention,'' in
  \emph{International Conference on Machine Learning}.\hskip 1em plus 0.5em
  minus 0.4em\relax PMLR, 2020, pp. 5156--5165.

\bibitem{karpushin2016keypoint}
M.~Karpushin, G.~Valenzise, and F.~Dufaux, ``Keypoint detection in rgbd images
  based on an anisotropic scale space,'' \emph{IEEE Transactions on
  Multimedia}, vol.~18, no.~9, pp. 1762--1771, 2016.

\bibitem{li2022exploiting}
W.~Li, H.~Liu, R.~Ding, M.~Liu, P.~Wang, and W.~Yang, ``Exploiting temporal
  contexts with strided transformer for 3d human pose estimation,'' \emph{IEEE
  Transactions on Multimedia}, 2022.

\bibitem{jiayao2022real}
S.~Jiayao, S.~Zhou, Y.~Cui, and Z.~Fang, ``Real-time 3d single object tracking
  with transformer,'' \emph{IEEE Transactions on Multimedia}, 2022.

\bibitem{pei2022transformer}
J.~Pei, T.~Cheng, H.~Tang, and C.~Chen, ``Transformer-based efficient salient
  instance segmentation networks with orientative query,'' \emph{IEEE
  Transactions on Multimedia}, 2022.

\bibitem{fu2018refinet}
K.~Fu, Q.~Zhao, and I.~Y.-H. Gu, ``Refinet: A deep segmentation assisted
  refinement network for salient object detection,'' \emph{IEEE Transactions on
  Multimedia}, vol.~21, no.~2, pp. 457--469, 2018.

\bibitem{song20166}
Y.~Song, X.~Chen, X.~Wang, Y.~Zhang, and J.~Li, ``6-dof image localization from
  massive geo-tagged reference images,'' \emph{IEEE Transactions on
  Multimedia}, vol.~18, no.~8, pp. 1542--1554, 2016.

\bibitem{cuturi2013sinkhorn}
M.~Cuturi, ``Sinkhorn distances: Lightspeed computation of optimal transport,''
  \emph{Advances in neural information processing systems}, vol.~26, 2013.

\bibitem{simonyan2014very}
K.~Simonyan and A.~Zisserman, ``Very deep convolutional networks for
  large-scale image recognition,'' \emph{arXiv preprint arXiv:1409.1556}, 2014.

\bibitem{tang2022quadtree}
S.~Tang, J.~Zhang, S.~Zhu, and P.~Tan, ``Quadtree attention for vision
  transformers,'' \emph{arXiv preprint arXiv:2201.02767}, 2022.

\bibitem{beltagy2020longformer}
I.~Beltagy, M.~E. Peters, and A.~Cohan, ``Longformer: The long-document
  transformer,'' \emph{arXiv preprint arXiv:2004.05150}, 2020.

\bibitem{wu2021fastformer}
C.~Wu, F.~Wu, T.~Qi, Y.~Huang, and X.~Xie, ``Fastformer: Additive attention can
  be all you need,'' \emph{arXiv preprint arXiv:2108.09084}, 2021.

\bibitem{he2016deep}
K.~He, X.~Zhang, S.~Ren, and J.~Sun, ``Deep residual learning for image
  recognition,'' in \emph{Proceedings of the IEEE conference on computer vision
  and pattern recognition}, 2016, pp. 770--778.

\bibitem{lin2017feature}
T.-Y. Lin, P.~Doll{\'a}r, R.~Girshick, K.~He, B.~Hariharan, and S.~Belongie,
  ``Feature pyramid networks for object detection,'' in \emph{Proceedings of
  the IEEE conference on computer vision and pattern recognition}, 2017, pp.
  2117--2125.

\bibitem{howard2017mobilenets}
A.~G. Howard, M.~Zhu, B.~Chen, D.~Kalenichenko, W.~Wang, T.~Weyand,
  M.~Andreetto, and H.~Adam, ``Mobilenets: Efficient convolutional neural
  networks for mobile vision applications,'' \emph{arXiv preprint
  arXiv:1704.04861}, 2017.

\bibitem{dai2017scannet}
A.~Dai, A.~X. Chang, M.~Savva, M.~Halber, T.~Funkhouser, and M.~Nie{\ss}ner,
  ``Scannet: Richly-annotated 3d reconstructions of indoor scenes,'' in
  \emph{Proceedings of the IEEE conference on computer vision and pattern
  recognition}, 2017, pp. 5828--5839.

\bibitem{loshchilov2017decoupled}
I.~Loshchilov and F.~Hutter, ``Decoupled weight decay regularization,''
  \emph{arXiv preprint arXiv:1711.05101}, 2017.

\bibitem{li2018megadepth}
Z.~Li and N.~Snavely, ``Megadepth: Learning single-view depth prediction from
  internet photos,'' in \emph{Proceedings of the IEEE Conference on Computer
  Vision and Pattern Recognition}, 2018, pp. 2041--2050.

\bibitem{balntas2017hpatches}
V.~Balntas, K.~Lenc, A.~Vedaldi, and K.~Mikolajczyk, ``Hpatches: A benchmark
  and evaluation of handcrafted and learned local descriptors,'' in
  \emph{Proceedings of the IEEE conference on computer vision and pattern
  recognition}, 2017, pp. 5173--5182.

\bibitem{cai2022htmatch}
Y.~Cai, L.~Li, D.~Wang, X.~Li, and X.~Liu, ``Htmatch: An efficient hybrid
  transformer based graph neural network for local feature matching,''
  \emph{Signal Processing}, p. 108859, 2022.

\end{thebibliography}
\end{document}